\documentclass{article}

\PassOptionsToPackage{square,numbers}{natbib}
\usepackage[preprint]{neurips_data_2023}

\usepackage[utf8]{inputenc} 
\usepackage[T1]{fontenc}    
\usepackage{hyperref}       
\usepackage{url}            
\usepackage{booktabs}       
\usepackage{amsfonts}       
\usepackage{nicefrac}       
\usepackage{microtype}      
\usepackage{xcolor} 
\usepackage{subcaption}     
\usepackage{amsmath}
\usepackage{enumitem}
\usepackage{authblk}
\usepackage{babel}

\usepackage{graphicx} 

\usepackage{listings}
\title{Open RL Benchmark: Comprehensive Tracked Experiments for Reinforcement Learning}

\author{
\textbf{Shengyi~Huang}$^{1,2*}$\quad
\textbf{Quentin~Gallouédec}$^3$\thanks{Equal contributions}\quad
\textbf{Florian~Felten}$^4$\quad
\textbf{Antonin~Raffin}$^5$\quad
\textbf{Rousslan~Fernand~Julien~Dossa}$^6$\quad
\textbf{Yanxiao~Zhao}$^{7,8}$\quad
\textbf{Ryan~Sullivan}$^9$\quad
\textbf{Viktor~Makoviychuk}$^{10}$\quad
\textbf{Denys~Makoviichuk}$^{11}$\quad
\textbf{Mohamad~H.~Danesh}$^{12}$\quad
\textbf{Cyril~Roumégous}$^{13}$\quad
\textbf{Jiayi~Weng}\quad
\textbf{Chufan~Chen}$^{14}$\quad
\textbf{Md~Masudur~Rahman}$^{15}$\quad
\textbf{João~G.~M.~Araújo}\thanks{Work done while at Cohere}\quad
\textbf{Guorui~Quan}$^{16}$\quad
\textbf{Daniel~Tan}$^{17,18}$\quad
\textbf{Timo~Klein}$^{19,20}$\quad
\textbf{Rujikorn~Charakorn}$^{21}$\quad
\textbf{Mark~Towers}$^{22}$\quad
\textbf{Yann~Berthelot}$^{23,24}$\quad
\textbf{Kinal~Mehta}$^{25}$\quad
\textbf{Dipam~Chakraborty}$^{26}$\quad
\textbf{Arjun~KG}\quad
\textbf{Valentin~Charraut}$^{27}$\quad
\textbf{Chang~Ye}$^{28}$\quad
\textbf{Zichen~Liu}$^{29}$\quad
\textbf{Lucas~N.~Alegre}$^{30}$\quad
\textbf{Alexander Nikulin}$^{31}$\quad
\textbf{Xiao Hu}$^{32}$\quad
\textbf{Tianlin Liu}$^{33}$\quad
\textbf{Jongwook~Choi}$^{34}$\quad
\textbf{Brent Yi}$^{35}$
}

\affil{\footnotesize{
$^{1}$Hugging Face\quad
$^{2}$Drexel University\quad
$^{3}$Univ. Lyon, Centrale Lyon, CNRS, INSA Lyon, UCBL, LIRIS, UMR 5205\quad
$^{4}$SnT, University of Luxembourg\quad
$^{5}$German Aerospace Center (DLR) RMC, Weßling, Germany\quad
$^{6}$Araya Inc., Tokyo, Japan\quad
$^{7}$School of Computer Science and Technology, University of Chinese Academy of Sciences\quad
$^{8}$Chengdu Institute of Computer Applications, Chinese Academy of Sciences\quad
$^{9}$University of Maryland, College Park\quad
$^{10}$NVIDIA\quad
$^{11}$Snap Inc.\quad
$^{12}$School of Computer Science, McGill University\quad
$^{13}$Polytech Montpellier DO\quad
$^{14}$Zhejiang University\quad
$^{15}$Department of Computer Science, Purdue University\quad
$^{16}$Chinese University of Hong Kong, Shenzhen\quad
$^{17}$University College London\quad
$^{18}$Agency for Science, Technology and Research\quad
$^{19}$Faculty of Computer Science, University of Vienna, Vienna, Austria\quad
$^{20}$UniVie Doctoral School Computer Science, University of Vienna\quad
$^{21}$Vidyasirimedhi Institute of Science and Technology (VISTEC)\quad
$^{22}$University of Southampton\quad
$^{23}$Univ. Lille, Inria, CNRS, Centrale Lille, UMR 9189 – CRIStAL\quad
$^{24}$Saint-Gobain Research Paris\quad
$^{25}$International Institute of Information Technology, Hyderabad, India\quad
$^{26}$AIcrowd SA\quad
$^{27}$Valeo Driving Assistance Research\quad
$^{28}$New York University\quad
$^{29}$Sea AI Lab\quad
$^{30}$Institute of Informatics, Federal University of Rio Grande do Sul\quad
$^{31}$Tinkoff\quad
$^{32}$Department of Automation, Tsinghua University\quad
$^{33}$University of Basel\quad
$^{34}$University of Michigan\quad
$^{35}$UC Berkeley
}}

\begin{document}

\maketitle

\begin{abstract}
In many Reinforcement Learning (RL) papers, learning curves are useful indicators to measure the effectiveness of RL algorithms. However, the complete raw data of the learning curves are rarely available. As a result, it is usually necessary to reproduce the experiments from scratch, which can be time-consuming and error-prone. 
We present Open RL Benchmark, a set of fully tracked RL experiments, including not only the usual data such as episodic return, but also all algorithm-specific and system metrics. Open RL Benchmark is community-driven: anyone can download, use, and contribute to the data. At the time of writing, more than 25,000 runs have been tracked, for a cumulative duration of more than 8 years. Open RL Benchmark covers a wide range of RL libraries and reference implementations. Special care is taken to ensure that each experiment is precisely reproducible by providing not only the full parameters, but also the versions of the dependencies used to generate it. In addition, Open RL Benchmark comes with a command-line interface (CLI) for easy fetching and generating figures to present the results. In this document, we include two case studies to demonstrate the usefulness of Open RL Benchmark in practice. To the best of our knowledge, Open RL Benchmark is the first RL benchmark of its kind, and the authors hope that it will improve and facilitate the work of researchers in the field.
\end{abstract}

\section{Introduction}
Reinforcement Learning (RL) research is based on comparing new methods to baselines to assess progress~\cite{patterson2023empirical}. 
This process implies the availability of the data associated with these baselines~\cite{raffin2021stable} or, alternatively, the ability to reproduce them and generate the data oneself~\cite{raffin2020rl}. 
In addition, the ability to reproduce also allows the methods to be compared with new benchmarks and to identify the areas in which the methods excel and those in which they are likely to fail, thus providing avenues for future research.

In practice, the RL research community faces complex challenges in comparing new methods with reference data. The unavailability of reference data requires researchers to reproduce experiments, posing difficulties due to insufficient source code documentation and evolving software dependencies. Implementation intricacies, as highlighted in past research, can significantly impact results \cite{henderson2018deep, huang202237}. Moreover, limited computing resources play a crucial role, hindering the reproduction process and affecting researchers without substantial access. These challenges lead to difficulties in reliably evaluating new methods and hinder efficient comparisons against established ones. Reproducing experiments is a time-consuming and resource-intensive task, or researchers may rely on inconsistently presented paper results. The lack of standardized metrics and benchmarks across studies not only impedes comparison but also results in a substantial waste of time and resources. To address these issues, the RL community must establish rigorous reproducibility standards, ensuring replicability and comparability across studies. Transparent sharing of data, code, and experimental details, along with the adoption of consistent metrics and benchmarks, would collectively enhance the evaluation and progression of RL research, ultimately accelerating advancements in the field.

Open RL Benchmark presents a rich collection of tracked RL experiments and aims to set a new standard by providing a diverse training dataset. This initiative prioritizes the use of existing data over re-running baselines, emphasizing reproducibility and transparency. Our contributions are:
\begin{itemize}[noitemsep,topsep=0pt,parsep=0pt,partopsep=0pt]
\item \textbf{Extensive dataset:} Offers a large, diverse collection of tracked RL experiments.
\item \textbf{Standardization:} Establishes a new norm by encouraging reliance on existing data, reducing the need for re-running baselines.
\item \textbf{Comprehensive metrics:} Includes diverse tracked metrics for method-specific and system evaluation, in addition to episodic return.
\item \textbf{Reproducibility:} Emphasizes clear instructions and fixed dependencies, ensuring easy experiment replication.
\item \textbf{Resource for research:} Serves as a valuable and collaborative resource for RL research.
\item \textbf{Facilitating exploration:} Enables reliable exploration and assessment of new RL methods.
\end{itemize}



\section{Comprehensive Overview of Open RL Benchmark: Content, Methodology, Tools, and Applications}

\begin{figure}
    \centering
    \includegraphics[width=\textwidth]{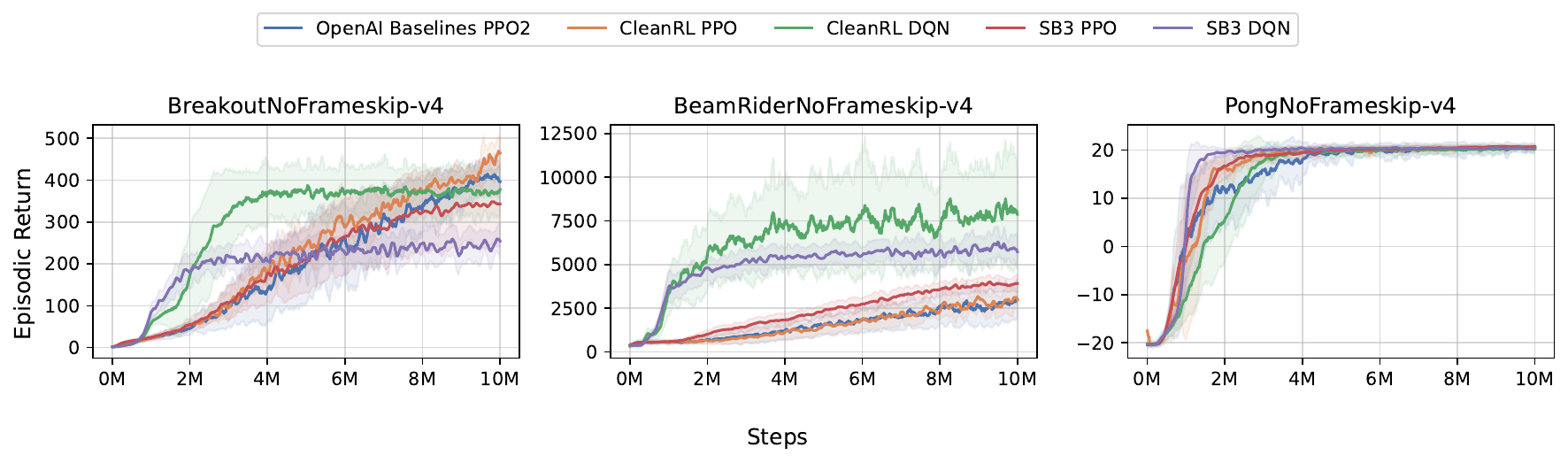}
    \caption{Example of learning curves obtained with Open RL Benchmark. These compare the episodic returns achieved by different implementations of PPO and DQN on a number of Atari games.}
    \label{fig:example_learning_curves}
\end{figure}

This section provides a detailed exploration of the contents of Open RL Benchmark, including its diverse set of libraries and environments, and the metrics it contains. We also look at the practical aspects of using Open RL Benchmark, highlighting its ability to ensure accurate reproducibility and facilitate the creation of data visualizations thanks to its CLI.

\subsection{Content}
\label{subsec:content}

\begin{figure}[ht]
    \centering
    \includegraphics[width=0.75\textwidth]{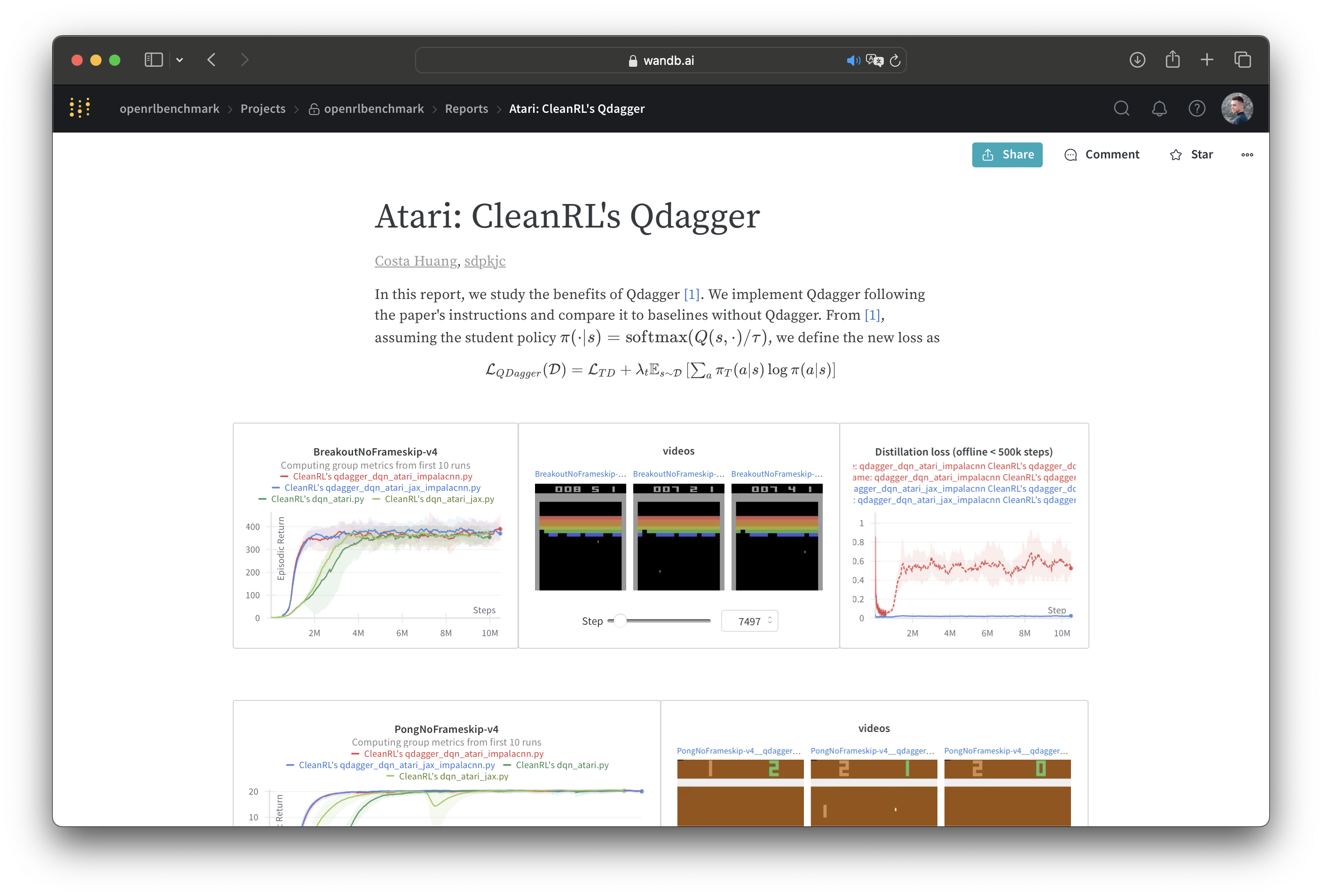}
    \caption{An example of a report on the Weights and Biases platform, dealing with the contribution of QDagger \cite{agarwal2022reincarnating}, and using data from Open RL Benchmark. The URL to access the report is \url{https://wandb.ai/openrlbenchmark/openrlbenchmark/reports/Atari-CleanRL-s-Qdagger--Vmlldzo0NTg1ODY5}}
    \label{fig:report}
\end{figure}

Open RL Benchmark data is stored and shared with Weights and Biases \cite{biewald2020experiment}. They are contained in a common entity named \texttt{openrlbenchmark}. Runs are divided into several \textit{projects}. A project can correspond to a library, but it can also correspond to a set of more specific runs, such as \texttt{envpool-cleanrl} in which we find CleanRL runs \cite{huang2022cleanrl} which have the particularity of being launched with the EnvPool implementation \cite{weng2022envpool} of environments. A project can also correspond to a reference implementation, such as TD3 (project \texttt{sfujim-TD3}) or Phasic Policy Gradient \cite{cobbe2021phasic} (project \texttt{phasic-policy-gradient}).
Open RL Benchmark also includes reports, which are interactive documents designed to enhance the visualization of selected representations. These reports provide a more user-friendly format for practitioners to share, discuss, and analyze experimental results, even across different projects. Figure \ref{fig:report} shows a preview of one such report.

At the time of writing, Open RL Benchmark contains nearly 25,000 runs, for a total of 72,000 hours (more than 8 years) of tracking.
In the following paragraphs, we present the libraries and environments for which runs are available in Open RL Benchmark, as well as the metrics tracked.

\paragraph{Libraries}
Open RL Benchmark contains runs for several reference RL libraries. These libraries are:
abcdRL \cite{zhao2022abcdrl},
Acme \cite{hoffman2020acme},
Cleanba \cite{huang2023cleanba},
CleanRL \cite{huang2022cleanrl},
jaxrl \cite{kostrikov2021jaxrl},
moolib \cite{mella2022moolib},
MORL-Baselines \cite{felten2023toolkit},
OpenAI Baselines \cite{dhariwal2017openai},
rlgames \cite{makoviichuk2021rl}
Stable Baselines3 \cite{raffin2021stable, raffin2020rl}
Stable Baselines Jax \cite{raffin2021stable} and 
TorchBeast \cite{k_uttler2019torchbeast}.

\paragraph{Environments}
The runs contained in Open RL Benchmark cover a wide range of classic environments. They include 
Atari \cite{bellemare2013arcade, machado2018revisiting},
Classic control \cite{brockman2016openai},
Box2d \cite{brockman2016openai} and 
MuJoCo \cite{todorov2012mujoco} as part of either Gym \cite{brockman2016openai} or
Gymnasium \cite{towers2023gymnasium} or EnvPool \cite{weng2022envpool}. They also include
Bullet \cite{coumans2016pybullet},
Procgen Benchmark \cite{cobbe2020leveraging},
Fetch environments \cite{plappert2018multi},
PandaGym \cite{gallouédec2021panda},
highway-env \cite{leurent2018environment},
Minigrid \cite{chevalier_boisvert2023minigrid} and
MO-Gymnasium \cite{alegre2022mo}.

\paragraph{Tracked metrics}

Metrics are recorded throughout the learning process, consistently linked with a global step indicating the number of interactions with the environment, and an absolute time, which allows for the calculation of the process's relative duration to track elapsed time. We categorize these metrics into four distinct groups:

\begin{itemize}[noitemsep,topsep=0pt,parsep=0pt,partopsep=0pt]
\item \textbf{Training-related metrics:} These are general metrics related to RL learning. This category contains, for example, the average returns obtained, the episode length or the number of collected samples per second.
\item \textbf{Method-specific metrics:} These are losses and measures of key internal values of the methods. For PPO, for example, this category includes the value loss, the policy loss, the entropy or the approximate KL divergence.
\item \textbf{Evolving configuration parameters:} These are configuration values that change during the learning process. This category includes, for example, the learning rate when there is decay, or the exploration rate ($\epsilon$) in the Deep Q-Network (DQN) \cite{mnih2013playing}.
\item \textbf{System metrics:} These are metrics related to system components. These could be GPU memory usage, its power consumption, its temperature, system and process memory usage, CPU usage or even network traffic.
\end{itemize}

The specific metrics available may vary from one library to another. In addition, even where the metrics are technically similar, the terminology or key used to record them may vary from one library to another. Users are advised to consult the documentation specific to each library for precise information on these measures.

\subsection{Everything you need for perfect repeatability}
\label{subsec:repeatability}

\begin{figure}[ht]
    \centering
    \includegraphics[width=\textwidth]{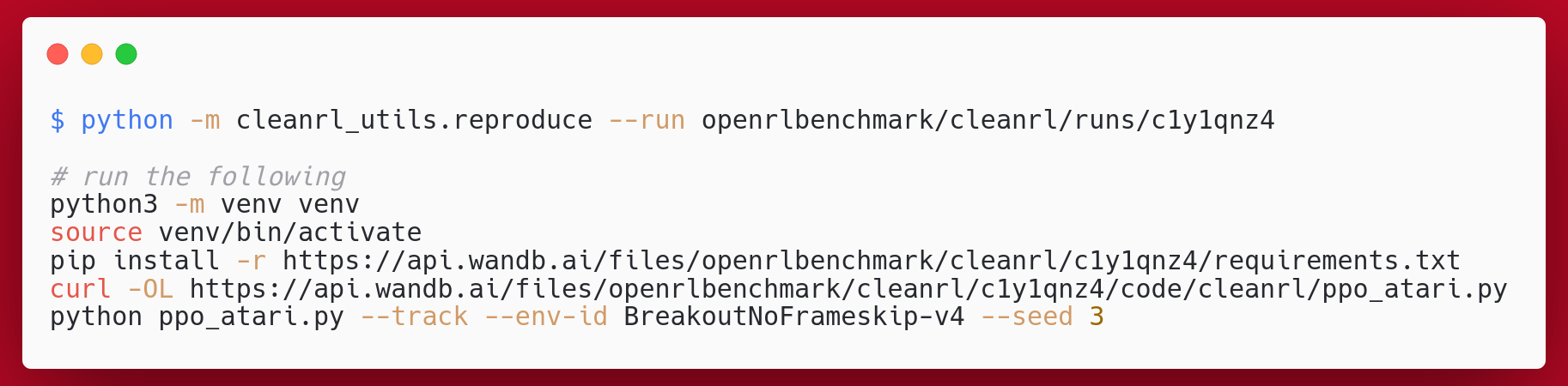}
    \caption{CleanRL's module \texttt{reproduce} allows the user to generate, from an Open RL Benchmark run reference, the exact command suite for an identical reproduction of the run.}
    \label{fig:repro}
\end{figure}

Reproducing experimental results in computational research, as discussed in Section \ref{subsec:reproducibility}, is often challenging due to evolving codebases, incomplete hyperparameter listings, version discrepancies, and compatibility issues. Our approach aims to enhance reproducibility by ensuring users can exactly replicate benchmark results. Each experiment includes a complete configuration with all hyperparameters, frozen versions of dependencies, and the exact command, including the necessary random seed, for systematic reproducibility.
Furthermore, CleanRL \cite{huang2022cleanrl} introduces a unique utility that streamlines the process of experiment replication (see Figure \ref{fig:repro}). This utility produces the command lines to set up a Python environment with the necessary dependencies, download the run file, and the precise command required for the experiment reproduction.
Such an approach to reproduction facilitates research and makes it possible to study in depth unusual phenomena, or cases of rupture\footnote{Exemplified in \url{https://github.com/DLR-RM/rl-baselines3-zoo/issues/427}}, in learning processes, which are generally ignored in the results presented, either because they are deliberately left out or because they are erased by the averaging process.

\subsection{The CLI, for figures in one command line}

Open RL Benchmark offers convenient access to raw data from RL libraries on standard environments. It includes a feature for easily extracting and visualizing data in a paper-friendly format, streamlining the process of filtering and extracting relevant runs and metrics for research papers through a single command. The CLI is a powerful tool for generating most metrics-related figures for RL research and notably, all figures in this document were generated using the CLI. The data in Open RL Benchmark can also be accessed by custom scripts, as detailed in Appendix~\ref{sec:appendix_custom_script}.
Specifically, the CLI integrated into Open RL Benchmark provides users with the flexibility to:

\begin{itemize}[noitemsep,topsep=0pt,parsep=0pt,partopsep=0pt]
\item Specify algorithms' implementations (from which library) along with their corresponding git commit or tag;
\item Choose target environments for analysis;
\item Define the metrics of interest;
\item Opt for the additional generation of metrics and plots using RLiable~\cite{agarwal2021deep}.
\end{itemize}

Concrete example usage of the CLI and resulting plots are available in Appendix~\ref{sec:appendix_cli}.

\section{Open RL Benchmark in Action: An Insight Into Case Studies}


Open RL Benchmark offers a powerful tool for researchers to evaluate and compare different RL algorithms. In this section, we'll explore two case studies that showcase its benefits.

First, we propose to investigate the effect of using TD($\lambda$) for value estimation in PPO \cite{schulman2017proximal} versus using Monte Carlo (MC). This simple study illustrates the use of Open RL Benchmark through a classic research question. Moreover, to the best of our knowledge, this question has never been studied in the literature.
We then present a more unusual approach. We show how Open RL Benchmark is used to demonstrate the speedup and variance reduction of a new IMPALA implementation proposed by \cite{huang2023cleanba}.


By using Open RL Benchmark, we can save time and resources while ensuring consistent and reproducible comparisons. These case studies highlight the role of the benchmark in providing insights that can advance the field of RL research.

\subsection{Easily assess the contribution of TD(\texorpdfstring{$\lambda$}{λ}) for value estimation in PPO}
\label{subsec:gae_for_ppo_value}

In the first case study, we show how Open RL Benchmark can be used to easily compare the performance of different methods for estimating the value function in PPO \cite{schulman2017proximal}, one of the many implementation details of this algorithm~\cite{huang202237}.
Specifically, we compare the commonly used Temporal Difference (TD)($\lambda$) estimate to the Monte-Carlo (MC) estimate.

PPO typically employs Generalized Advantage Estimation (GAE, \cite{schulman2016high}) to update the actor.
The advantage estimate is expressed as follows:

\begin{align}
    A^{\mathrm{GAE}(\gamma,\lambda)}_t = \sum_{l=0}^{N-1} (\gamma \lambda)^l \delta_{t+l}^V
\end{align}
where $\lambda\in[0,1]$ adjusts the bias-variance tradeoff and $\delta_{t+l}^V = R_{t+l} + \gamma \hat{V}(S_{t+l+1}) - \hat{V}(S_{t+l})$.

The target return for critic optimization is estimated with TD($\lambda$) as follows:

\begin{align}
  G_t^\lambda = (1-\lambda)\sum_{n=1}^\infty \lambda^{n-1}G_{t:t+n}
\end{align}

where $G_{t:t+n} = \sum_{k=0}^{n-1}\gamma^k R_{t+k+1}+\gamma^nV(S_{t+n})$ is the $n$-steps return.

In practice, the target return for updating the critic is computed from the GAE value, by adding the minibatch return, a detail usually overlooked by practitioners \citep[point 5]{huang202237}.
While previous studies~\cite{patterson2023empirical} have shown the joint benefit of GAE and TD($\lambda$) over MC estimates for actor and critic, we focus on the value function alone.
To isolate the influence of the value function estimation, we vary the method used for the value function and keep GAE for advantage estimation.

\begin{figure}[ht]
    \centering
    \subcaptionbox{Results for Atari games}{
        \includegraphics[width=0.47\textwidth]{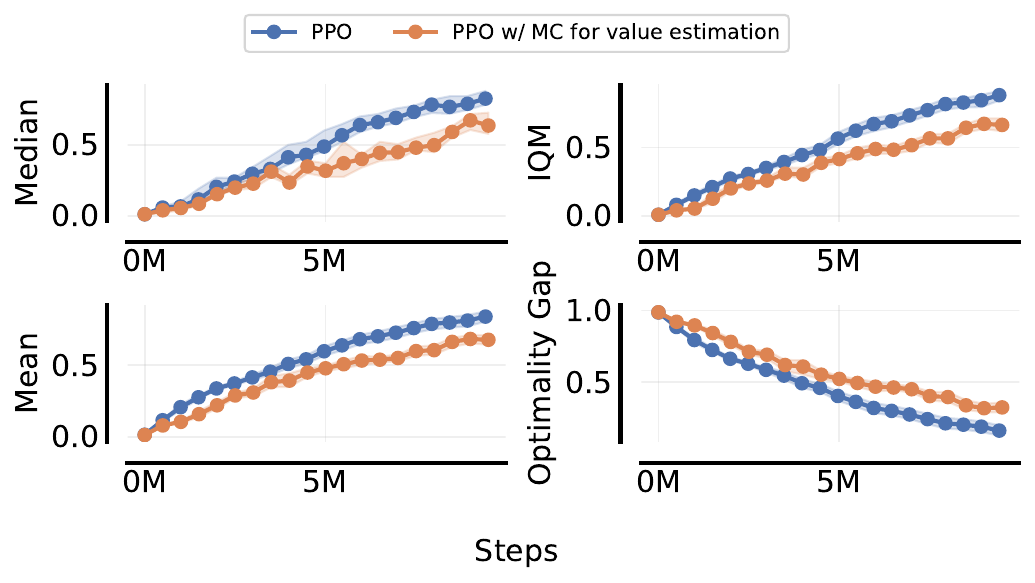}
    }
    \subcaptionbox{Results for Box2D and MuJoCo environments}{
        \includegraphics[width=0.47\textwidth]{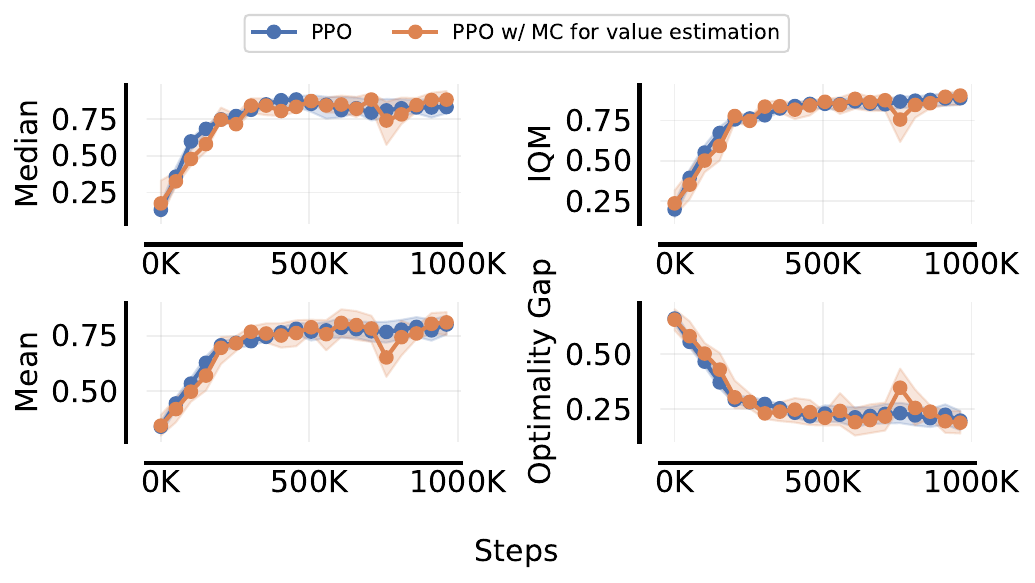}
    }
    \caption{Comparing the original PPO and the PPO with Monte-Carlo (MC) for value estimation. These experiments were conducted over 15 environments, including Atari games, Box2D, and MuJoCo. Plot shows minmax normalized scores with 95\% stratified bootstrap CIs.}
    \label{fig:compare_ppo_sample_efficiency}
\end{figure}

The first step is to identify the reference runs in Open RL Benchmark. As PPO is a widely recognized baseline, a large number of runs are available. We chose to use the Stable Baselines3 runs for this example. We retrieve the precise source code and command used to generate them, thanks to the pinned dependencies provided in the runs. We apply the appropriate modification to the source code. For each environment selected, we launch 3 learning runs using the same command as the one retrieved. The runs are stored in a dedicated project \footnote{\url{https://wandb.ai/modanesh/openrlbenchmark}}. For fast and user-friendly rendering of the results, we create a Weights and Biases report\footnote{\url{https://api.wandb.ai/links/modanesh/izf4yje4}}. Using the CLI, we generate Figure \ref{fig:compare_ppo_sample_efficiency} and \ref{fig:compare_ppo_aggregate}. The command used to generate the figures is given in Appendix \ref{sec:appendix_detailed}.

Figure \ref{fig:compare_ppo_sample_efficiency} gives an overview of the results, while detailed plots in the Appendix~\ref{sec:appendix_detailed} provide a closer look at each environment.
The proposed modification to the PPO value function estimation does have an impact on its performance and efficiency.
These experiments were run over various environments including Atari games (Breakout, Space Invaders, Seaquest, Enduro, Pong, Q*Bert, Beam Rider), Box2D (Lunar Lander), and MuJoCo (Inverted Double Pendulum, Inverted Pendulum, Reacher, Half Cheetah, Hopper, Swimmer, Walker 2d). Results presented in Figure \ref{fig:compare_ppo_sample_efficiency} serve as powerful tools for elucidating the nuanced dynamics of the modified approach, offering a clear and intuitive portrayal of the impact on learning dynamics and convergence rates. This would provide a very valuable tool for researchers to navigate through various ideas, and uncover patterns and distinctions that not only validate the efficacy of their changes but also contribute to a deeper understanding of the underlying mechanisms at play. As an example, from Figure \ref{fig:compare_ppo_sample_efficiency} (a), one could observe that the proposed idea resulted in a worse performance compared to the original PPO on Atari games. On the other hand, from Figure \ref{fig:compare_ppo_sample_efficiency} (b), the performance of the proposed PPO is on par with the original PPO in Box2D and MuJoCo environments.

Figure \ref{fig:compare_ppo_aggregate} demonstrates that PPO maintains uniform learning metrics across evaluations, while PPO with MC for value estimation shows more variability in performance measures, even when the same hyperparameters are used.

\begin{figure}[ht]
    \centering
    \subcaptionbox{Results for Atari games}{
        \includegraphics[width=0.96\textwidth]{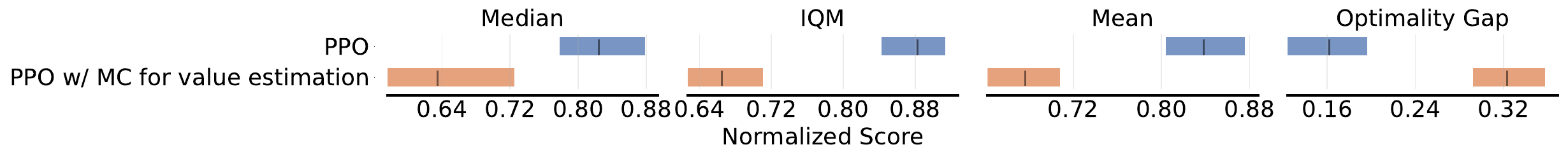}
    }
    \subcaptionbox{Results for Box2D and MuJoCo environments}{
        \includegraphics[width=0.96\textwidth]{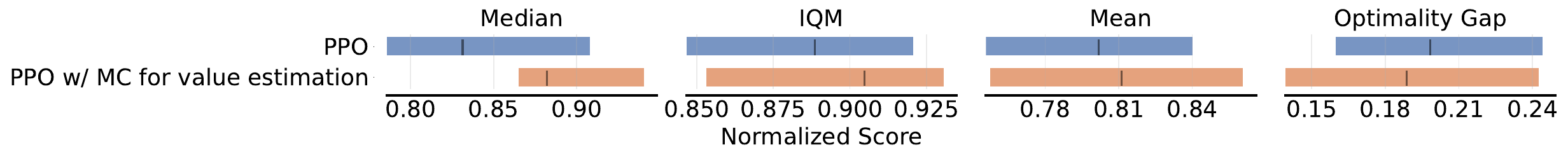}
    }
    \caption{Study of the contribution of GAE for estimating the value used to update the critic in PPO, compared against its variant which uses the MC estimator instead. Figures show the aggregated min-max normalized scores with stratified 95\% stratified bootstrap CIs.}
    \label{fig:compare_ppo_aggregate}
\end{figure}

\subsection{Demonstrating the utility of Open RL Benchmark through the Cleanba case study}

This section describes how Open RL Benchmark was instrumental in the evaluation and presentation of Cleanba \cite{huang2023cleanba}, a new open-source platform for distributed RL implementing highly optimized distributed variants of PPO \cite{schulman2017proximal} and IMPALA \cite{espeholt2018impala}.
Cleanba's authors asserted three points: (1) Cleanba implementations compare favorably with baselines in terms of sample efficiency, (2) for the same system, the Cleanba implementation is more optimized and therefore faster, and (3) the design choices allow a reduction in the variability of results.

To prove these assertions, the evaluation of Cleanba encountered a common problem in RL research: the works that initially proposed these baselines did not provide the raw results of their experiments. Although a reference implementation is available\footnote{\url{https://github.com/google-deepmind/scalable_agent}}, it is no longer maintained. Subsequent works such as Moolib \cite{mella2022moolib} and TorchBeast \cite{k_uttler2019torchbeast} have successfully replicated the IMPALA results. However, these shared results are limited to the paper's presented curves, which provide a smoothed measure of episodic return as a function of interaction steps on a specific set of Atari tasks. It's worth noting that these tasks are not an exact match for the widely recognized Atari 57, and the raw data used to generate these curves is unavailable.

Recognizing the lack of raw data for existing IMPALA implementations, the authors reproduced the experiments, tracked the runs and integrated them into Open RL Benchmark. As a reminder, these logged data include not only the return curves, but also the system configurations and temporal data, which are crucial to support the Cleanba authors' optimization claim. Comparable experiments have been run, tracked and shared on Open RL Benchmark with the proposed Cleanba implementation.

\begin{figure}[ht]
    \centering
    \includegraphics[width=0.8\textwidth]{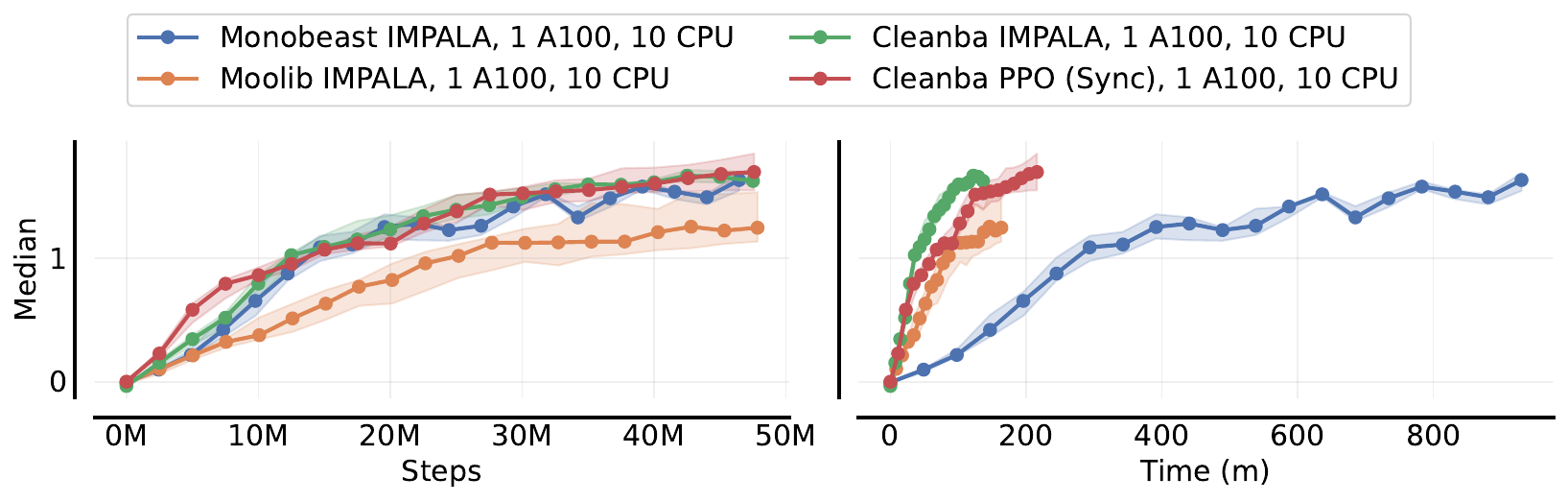}
    \caption{Median human-normalized scores with 95\% stratified bootstrap CIs of Cleanba \cite{huang2023cleanba} variants compared with moolib \cite{mella2022moolib} and monobeast \cite{k_uttler2019torchbeast}. The experiments were conducted on 57 Atari games \cite{bellemare2013arcade}. The data used to generate the figure comes from Open RL Benchmark, and the figure was generated with a single command from Open RL Benchmark's CLI. Figure from \cite{huang2023cleanba}.}
    \label{fig:cleanba1}
\end{figure}

\begin{figure}[ht]
    \centering
    \includegraphics[width=\textwidth]{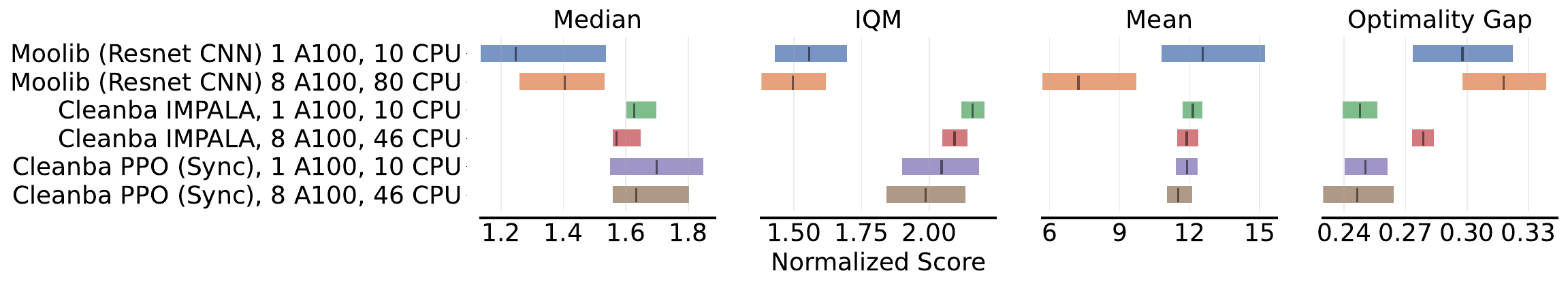}
    \caption{Aggregated normalized human scores with stratified 95\% bootstrap CIs, showing that unlike moolib \cite{mella2022moolib}, Cleanba \cite{huang2023cleanba} variants have more predictable learning curves (using the same hyperparameters) across different hardware configurations. Figure from \cite{huang2023cleanba}.}
    \label{fig:cleanba2}
\end{figure}

Using Open RL Benchmark CLI, the authors generated several figures. The authors have provided the exact commands to reproduce these curves in the source directory. In Figure \ref{fig:cleanba1}, taken from \cite{huang2023cleanba}, the authors show that the results in terms of sample efficiency compare favourably with the baselines, and that for the same system configuration, convergence was temporally faster with the proposed implementation, thus proving claims (1) and (2).
Figure \ref{fig:cleanba2} demonstrates that Cleanba variants maintain consistent learning curves across different hardware configurations. Conversely, moolib's IMPALA shows marked variability in similar settings, despite identical hyperparameters, affirming the authors' third claim.

\section{Current Practices in RL: Data Reporting, Sharing and Reproducibility}

Many new methods have emerged in recent years, with some becoming standard baselines, but current practices in the field make it challenging to interpret, compare, and replicate study results.
In this section, we highlight the inconsistent presentation of results, focusing on learning curves as an example. This inconsistency can hinder interpretation and lead to incorrect conclusions. We also note the insufficient availability of learning data, despite some positive efforts, and examine challenges related to method reproducibility.

\subsection{Analyzing learning curve practices}

Plotting learning curves is a common way to show the evolution of an agent's performance as it learns.
In this section, we take a closer look at the different components of learning curves. We examine in detail the choices made by a selection of key publications in the field on these different aspects. We show that among these publications, there is no uniformity on any aspect, that the choices of presentation are almost never motivated and that, sometimes, they are not even explicitly stated.

\paragraph{Axis}
Typically, the $y$ axis measures either the return acquired during data collection or evaluation. Some older papers, like \cite{schulman2015trust, mnih2016asynchronous, schulman2017proximal}, fail to specify the metric, using the vague term \textit{learning curve}. The first approach sums the rewards collected during agent rollout \cite{dabney2018implicit, burda2019exploration}). The second approach suspends training, averaging the agent's return over episodes, deactivating exploration elements \cite{fujimoto2018addressing, haarnoja2018soft, hessel2018rainbow, janner2019when, badia2020never, ecoffet2021first, chen2021randomized}.
This method is prevalent and provides a more precise evaluation. Regarding the $x$ axis, while older baselines \cite{schulman2015trust, mnih2016asynchronous} use policy updates and learning epochs, the norm is to use interaction counts with the environment. In Atari environments, it's often the number of frames, adjusting for frame skipping to match human interaction frequency.

\paragraph{Shaded area}
Data variability is typically shown with a shaded area, but its definition varies across studies. Commonly, it represents the standard deviation \cite{chen2021randomized, janner2019when} and less commonly half the standard deviation \cite{fujimoto2018addressing}. \cite{haarnoja2018soft} uses a min-max representation to include outliers, covering the entire observed range. This method offers a comprehensive view but amplifies outliers' impact with more runs. \cite{ecoffet2021first} adopts a probabilistic approach, showing a 95\% bootstrap confidence interval around the mean, ensuring statistical confidence. Unfortunately, \cite{schulman2015trust, schulman2017proximal, mnih2016asynchronous, dabney2018implicit, badia2020never} omit statistical details or even the shaded area, introducing uncertainty in data variability interpretation, as seen in \cite{hessel2018rainbow}.

\paragraph{Smoothing}
The variability in results can hinder figure clarity. While many papers present raw curves \cite{schulman2015trust, fujimoto2018addressing, haarnoja2018soft, dabney2018implicit, janner2019when, chen2021randomized, badia2020never}, smoothing is a common practice to address this issue. However, smoothing sacrifices data variability information. Authors should provide clear explanations of this post-treatment to prevent misinterpretation. Unfortunately, most authors don't offer sufficient details to understand and reproduce the smoothing process. For instance, \cite{schulman2015trust, schulman2017proximal} likely use smoothing without explicit mention. \cite{hessel2018rainbow, fujimoto2018addressing} briefly mention curve smoothing but lack method details. The exception is \cite{ecoffet2021first}, which provides a precise smoothing description.

\paragraph{Normalization and aggregation}
Performance aggregation assesses method results across various tasks and domains, indicating their generality and robustness.
Outside the Atari context, aggregation practices are not common due to the absence of a universal normalization standard.
In the absence of a widely accepted normalization strategy, scores are typically not aggregated, or if done, it relies on a min-max approach based on extreme study scores, lacking absolute significance and unsuitable for subsequent comparisons.
In the case of Atari, early research did not use normalization or aggregate results \cite{mnih2013playing}. However, there has been a notable shift towards generalizing normalization against human performance, although this has weaknesses and may not truly reflect agent mastery \cite{toromanoff2019is}. 
Aggregation methods vary as well. Mean is common but can be influenced by outliers, leading some studies to prefer the more robust median, as in \cite{hessel2018rainbow}, while, many papers now report both mean and median results \cite{dabney2018implicit, hafner2023mastering, badia2020agent57}. Recent approaches like \cite{lee2022multi} use the Interquartile Mean (IQM) for balanced aggregation, providing a more accurate performance representation across diverse games, as suggested by \cite{agarwal2021deep}.

\subsection{Spectrum of data sharing practices}

While the mentioned studies often have reference implementations (see Section \ref{subsec:reproducibility}), the sharing of training data typically extends only to the curves presented in their articles. This necessitates reliance on libraries that replicate these methods, offering benchmarks with varying levels of completeness.
Several widely-used libraries in the field provide high-level summaries or graphical representations without including raw data (e.g., Tensorforce \cite{kuhnle2017tensorforce}, Garage \cite{contributors2019garage}, ACME \cite{hoffman2020acme}, MushroomRL \cite{d_eramo2021mushroomrl}, ChainerRL \cite{fujita2021chainerrl}, and TorchRL \cite{bou2023torchrl}).
Spinning Up \cite{achiam2018spinning} offers partial data accessibility, providing benchmark curves but withholding raw data. TF-Agent \cite{guadarrama2018tf} is slightly better, offering experiment tracking with links to TensorBoard.dev, though its future is uncertain due to service closure.
Tianshou \cite{weng2022tianshou} provides individual run reward data for Atari and average rewards for MuJoCo, with more detailed MuJoCo data available via a Google Drive link, but it's not widely promoted.
RLLib \cite{liang2018rllib} maintains an intermediate stance in data sharing, hosting run data in a dedicated repository. However, this data is specific to select experiments and often presented in non-standard, undocumented formats, complicating its use.
Leading effective data-sharing platforms include Dopamine \cite{castro2018dopamine} and Sample Factory \cite{petrenko2020sample}. Dopamine consistently provides accessible raw evaluation data for various seeds and visualizations, along with trained agents on Google Cloud. Sample Factory offers comprehensive data via Weights and Biases \cite{biewald2020experiment} and a selection of pre-trained agents on the Hugging Face Hub, enhancing reproducibility and collaborative research efforts.

\subsection{Review on reproducibility}
\label{subsec:reproducibility}

The literature shows variations in these practices. Some older publications like \cite{schulman2015trust, schulman2017proximal, bellemare2013arcade, mnih2016asynchronous, hessel2018rainbow} and even recent ones like \cite{reed2022generalist} lack a codebase but provide detailed descriptions for replication\footnote{This section uses the taxonomy introduced by \cite{lynnerup2019survey}: \textit{repeatability} means accurately duplicating an experiment with source code and random seed availability, \textit{reproducibility} involves redoing an experiment using an existing codebase, and \textit{replicability} aims to achieve similar results independently through algorithm implementation.}. However, challenges arise because certain hyperparameters, important but often unreported, can significantly affect performance \cite{andrychowicz2020what}. In addition, implementation choices have proven to be critical \cite{henderson2018deep, huang2023cleanba, huang202237, engstrom2020implementation}, complicating the distinction between implementation-based improvements and methodological advances.

Recognizing these challenges, the RL community is advocating for higher standards. NeurIPS, for instance, has been requesting a reproduction checklist since 2019 \cite{pineau2021improving}. Recent efforts focus on systematic sharing of source code to promote reproducibility. However, codebases are often left unmaintained post-publication (with rare exceptions \cite{fujimoto2018addressing}), creating complexity for users dealing with various dependencies and unsolved issues. 
To address these challenges, libraries have aggregated multiple baseline implementations (see Section \ref{subsec:content}), aiming to match reported paper performance. However, long-term sustainability remains a concern. While these libraries enhance reproducibility, in-depth repeatability is still rare.

\section{Discussion and Conclusion}

Reproducing RL results is difficult due to limited data access and code sharing. Minor implementation variations can lead to performance differences, and verifying implementations lacks tools. Researchers often rely on vague comparisons with paper figures, making reproduction time-consuming and challenging, highlighting reliability and reproducibility issues in RL research. In our paper, we introduce Open RL Benchmark, a vast collection of tracked experiments spanning algorithms, libraries, and benchmarks. We capture all relevant metrics and data points, offering detailed resources for precise reproduction. This tool democratizes access to comprehensive datasets, simplifying valuable information extraction, enabling metric comparisons, and introducing a CLI for easier data access and visualization. Open RL Benchmark is a dynamic resource, regularly updated by both its founders and the user community. User contributions, whether new results or additional runs, enhance result reliability. Sharing trained agents can also offer insights and support offline RL studies.


Despite its strengths, Open RL Benchmark faces challenges in user-friendliness which must be addressed. Inconsistencies across libraries in evaluation strategies and terminology can complicate usage. Scaling community engagement becomes challenging with more members, libraries, and runs. The lack of Git-like version tracking for runs adds to these limitations. 

Open RL Benchmark is a key step forward in addressing RL research challenges. It offers a comprehensive, accessible, and collaborative experiment database, enabling precise comparisons and analyses. It enhances data access, promoting a deeper understanding of algorithmic performance. While challenges persist, Open RL Benchmark has the potential to elevate RL research standards.

\section*{Acknowledgements}

This work has been supported by a highly committed RL community. We have listed all the contributors to date, and would like to thank all future contributors and users in advance. 

This work was granted access to the HPC resources of IDRIS under the allocation 2022-[AD011012172R1] made by GENCI. 
The MORL-Baselines experiments have been conducted on the HPCs of the University of Luxembourg, and of the Vrije Universiteit Brussel.
This work was partly supported by the National Key Research and Development Program of China (2023YFB3308601), Science and Technology Service Network Initiative (KFJ-STS-QYZD-2021-21-001), the Talents by Sichuan provincial Party Committee Organization Department, and Chengdu - Chinese Academy of Sciences Science and Technology Cooperation Fund Project (Major Scientific and Technological Innovation Projects). Some experiments are conducted at Stability AI and Hugging Face's cluster. 

\bibliography{references}

\begin{thebibliography}{67}
\providecommand{\natexlab}[1]{#1}
\providecommand{\url}[1]{\texttt{#1}}
\expandafter\ifx\csname urlstyle\endcsname\relax
  \providecommand{\doi}[1]{doi: #1}\else
  \providecommand{\doi}{doi: \begingroup \urlstyle{rm}\Url}\fi

\bibitem[Achiam(2018)]{achiam2018spinning}
J.~Achiam.
\newblock {Spinning Up in Deep Reinforcement Learning}.
\newblock \url{https://github.com/openai/spinningup}, 2018.
\newblock URL \url{https://github.com/openai/spinningup}.

\bibitem[Agarwal et~al.(2021)Agarwal, Schwarzer, Castro, Courville, and Bellemare]{agarwal2021deep}
R.~Agarwal, M.~Schwarzer, P.~S. Castro, A.~C. Courville, and M.~G. Bellemare.
\newblock {Deep Reinforcement Learning at the Edge of the Statistical Precipice}.
\newblock In M.~Ranzato, A.~Beygelzimer, Y.~N. Dauphin, P.~Liang, and J.~W. Vaughan, editors, \emph{Advances in Neural Information Processing Systems 34: Annual Conference on Neural Information Processing Systems 2021, NeurIPS 2021, December 6-14, 2021, virtual}, pages 29304--29320, 2021.

\bibitem[Agarwal et~al.(2022)Agarwal, Schwarzer, Castro, Courville, and Bellemare]{agarwal2022reincarnating}
R.~Agarwal, M.~Schwarzer, P.~S. Castro, A.~C. Courville, and M.~G. Bellemare.
\newblock {Reincarnating Reinforcement Learning: Reusing Prior Computation to Accelerate Progress}.
\newblock In S.~Koyejo, S.~Mohamed, A.~Agarwal, D.~Belgrave, K.~Cho, and A.~Oh, editors, \emph{Advances in Neural Information Processing Systems 35: Annual Conference on Neural Information Processing Systems 2022, NeurIPS 2022, New Orleans, LA, USA, November 28 - December 9, 2022}, 2022.
\newblock URL \url{http://papers.nips.cc/paper\_files/paper/2022/hash/ba1c5356d9164bb64c446a4b690226b0-Abstract-Conference.html}.

\bibitem[Alegre et~al.(2022)Alegre, Felten, Talbi, Danoy, Now{\'e}, Bazzan, and da~Silva]{alegre2022mo}
L.~N. Alegre, F.~Felten, E.-G. Talbi, G.~Danoy, A.~Now{\'e}, A.~L.~C. Bazzan, and B.~C. da~Silva.
\newblock {MO-Gym: A Library of Multi-Objective Reinforcement Learning Environments}.
\newblock In \emph{Proceedings of the 34th Benelux Conference on Artificial Intelligence BNAIC/Benelearn 2022}, 2022.

\bibitem[Andrychowicz et~al.(2020)Andrychowicz, Raichuk, Stanczyk, Orsini, Girgin, Marinier, Hussenot, Geist, Pietquin, Michalski, Gelly, and Bachem]{andrychowicz2020what}
M.~Andrychowicz, A.~Raichuk, P.~Stanczyk, M.~Orsini, S.~Girgin, R.~Marinier, L.~Hussenot, M.~Geist, O.~Pietquin, M.~Michalski, S.~Gelly, and O.~Bachem.
\newblock {What Matters In On-Policy Reinforcement Learning? A Large-Scale Empirical Study}.
\newblock \emph{arXiv preprint arXiv:2006.05990}, 2020.

\bibitem[Badia et~al.(2020{\natexlab{a}})Badia, Piot, Kapturowski, Sprechmann, Vitvitskyi, Guo, and Blundell]{badia2020agent57}
A.~P. Badia, B.~Piot, S.~Kapturowski, P.~Sprechmann, A.~Vitvitskyi, Z.~D. Guo, and C.~Blundell.
\newblock {Agent57: Outperforming the Atari Human Benchmark}.
\newblock In \emph{Proceedings of the 37th International Conference on Machine Learning, {ICML} 2020, 13-18 July 2020, Virtual Event}, volume 119 of \emph{Proceedings of Machine Learning Research}, pages 507--517. {PMLR}, 2020{\natexlab{a}}.
\newblock URL \url{http://proceedings.mlr.press/v119/badia20a.html}.

\bibitem[Badia et~al.(2020{\natexlab{b}})Badia, Sprechmann, Vitvitskyi, Guo, Piot, Kapturowski, Tieleman, Arjovsky, Pritzel, Bolt, and Blundell]{badia2020never}
A.~P. Badia, P.~Sprechmann, A.~Vitvitskyi, Z.~D. Guo, B.~Piot, S.~Kapturowski, O.~Tieleman, M.~Arjovsky, A.~Pritzel, A.~Bolt, and C.~Blundell.
\newblock {Never Give Up: Learning Directed Exploration Strategies}.
\newblock In \emph{8th International Conference on Learning Representations, {ICLR} 2020, Addis Ababa, Ethiopia, April 26-30, 2020}. OpenReview.net, 2020{\natexlab{b}}.
\newblock URL \url{https://openreview.net/forum?id=Sye57xStvB}.

\bibitem[Bellemare et~al.(2013)Bellemare, Naddaf, Veness, and Bowling]{bellemare2013arcade}
M.~G. Bellemare, Y.~Naddaf, J.~Veness, and M.~Bowling.
\newblock {The Arcade Learning Environment: An Evaluation Platform for General Agents}.
\newblock \emph{Journal of Artificial Intelligence Research}, 47:\penalty0 253--279, 2013.
\newblock \doi{10.1613/JAIR.3912}.
\newblock URL \url{https://doi.org/10.1613/jair.3912}.

\bibitem[Biewald(2020)]{biewald2020experiment}
L.~Biewald.
\newblock {Experiment Tracking with Weights and Biases}, 2020.
\newblock URL \url{https://www.wandb.com/}.
\newblock Software available from wandb.com.

\bibitem[Bou et~al.(2023)Bou, Bettini, Dittert, Kumar, Sodhani, Yang, Fabritiis, and Moens]{bou2023torchrl}
A.~Bou, M.~Bettini, S.~Dittert, V.~Kumar, S.~Sodhani, X.~Yang, G.~D. Fabritiis, and V.~Moens.
\newblock {TorchRL: A Data-Driven Decision-Making Library for Pytorch}.
\newblock \emph{arXiv preprint arXiv:2306.00577}, 2023.

\bibitem[Brockman et~al.(2016)Brockman, Cheung, Pettersson, Schneider, Schulman, Tang, and Zaremba]{brockman2016openai}
G.~Brockman, V.~Cheung, L.~Pettersson, J.~Schneider, J.~Schulman, J.~Tang, and W.~Zaremba.
\newblock {OpenAI Gym}.
\newblock \emph{arXiv preprint arXiv:1606.01540}, 2016.

\bibitem[Burda et~al.(2019)Burda, Edwards, Storkey, and Klimov]{burda2019exploration}
Y.~Burda, H.~Edwards, A.~J. Storkey, and O.~Klimov.
\newblock {Exploration by random network distillation}.
\newblock In \emph{7th International Conference on Learning Representations, {ICLR} 2019, New Orleans, LA, USA, May 6-9, 2019}. OpenReview.net, 2019.
\newblock URL \url{https://openreview.net/forum?id=H1lJJnR5Ym}.

\bibitem[Castro et~al.(2018)Castro, Moitra, Gelada, Kumar, and Bellemare]{castro2018dopamine}
P.~S. Castro, S.~Moitra, C.~Gelada, S.~Kumar, and M.~G. Bellemare.
\newblock {Dopamine: A Research Framework for Deep Reinforcement Learning}.
\newblock \emph{arXiv preprint arXiv:1812.06110}, 2018.

\bibitem[Chen et~al.(2021)Chen, Wang, Zhou, and Ross]{chen2021randomized}
X.~Chen, C.~Wang, Z.~Zhou, and K.~W. Ross.
\newblock {Randomized Ensembled Double Q-Learning: Learning Fast Without a Model}.
\newblock In \emph{9th International Conference on Learning Representations, {ICLR} 2021, Virtual Event, Austria, May 3-7, 2021}. OpenReview.net, 2021.
\newblock URL \url{https://openreview.net/forum?id=AY8zfZm0tDd}.

\bibitem[Chevalier-Boisvert et~al.(2023)Chevalier-Boisvert, Dai, Towers, de~Lazcano, Willems, Lahlou, Pal, Castro, and Terry]{chevalier_boisvert2023minigrid}
M.~Chevalier-Boisvert, B.~Dai, M.~Towers, R.~de~Lazcano, L.~Willems, S.~Lahlou, S.~Pal, P.~S. Castro, and J.~Terry.
\newblock {Minigrid \& Miniworld: Modular \& Customizable Reinforcement Learning Environments for Goal-Oriented Tasks}.
\newblock \emph{arXiv preprint arXiv:2306.13831}, 2023.

\bibitem[Cobbe et~al.(2020)Cobbe, Hesse, Hilton, and Schulman]{cobbe2020leveraging}
K.~Cobbe, C.~Hesse, J.~Hilton, and J.~Schulman.
\newblock {Leveraging Procedural Generation to Benchmark Reinforcement Learning}.
\newblock In \emph{Proceedings of the 37th International Conference on Machine Learning, {ICML} 2020, 13-18 July 2020, Virtual Event}, volume 119 of \emph{Proceedings of Machine Learning Research}, pages 2048--2056. {PMLR}, 2020.
\newblock URL \url{http://proceedings.mlr.press/v119/cobbe20a.html}.

\bibitem[Cobbe et~al.(2021)Cobbe, Hilton, Klimov, and Schulman]{cobbe2021phasic}
K.~Cobbe, J.~Hilton, O.~Klimov, and J.~Schulman.
\newblock {Phasic Policy Gradient}.
\newblock In M.~Meila and T.~Zhang, editors, \emph{Proceedings of the 38th International Conference on Machine Learning, {ICML} 2021, 18-24 July 2021, Virtual Event}, volume 139 of \emph{Proceedings of Machine Learning Research}, pages 2020--2027. {PMLR}, 2021.
\newblock URL \url{http://proceedings.mlr.press/v139/cobbe21a.html}.

\bibitem[Coumans and Bai(2016)]{coumans2016pybullet}
E.~Coumans and Y.~Bai.
\newblock {PyBullet, a Python Module for Physics Simulation for Games, Robotics and Machine Learning}.
\newblock 2016.

\bibitem[Dabney et~al.(2018)Dabney, Ostrovski, Silver, and Munos]{dabney2018implicit}
W.~Dabney, G.~Ostrovski, D.~Silver, and R.~Munos.
\newblock {Implicit Quantile Networks for Distributional Reinforcement Learning}.
\newblock In J.~G. Dy and A.~Krause, editors, \emph{Proceedings of the 35th International Conference on Machine Learning, {ICML} 2018, Stockholmsm{\"{a}}ssan, Stockholm, Sweden, July 10-15, 2018}, volume~80 of \emph{Proceedings of Machine Learning Research}, pages 1104--1113. {PMLR}, 2018.
\newblock URL \url{http://proceedings.mlr.press/v80/dabney18a.html}.

\bibitem[D'Eramo et~al.(2021)D'Eramo, Tateo, Bonarini, Restelli, and Peters]{d_eramo2021mushroomrl}
C.~D'Eramo, D.~Tateo, A.~Bonarini, M.~Restelli, and J.~Peters.
\newblock {MushroomRL: Simplifying Reinforcement Learning Research}.
\newblock \emph{Journal of Machine Learning Research}, 22\penalty0 (131):\penalty0 1--5, 2021.
\newblock URL \url{http://jmlr.org/papers/v22/18-056.html}.

\bibitem[Dhariwal et~al.(2017)Dhariwal, Hesse, Klimov, Nichol, Plappert, Radford, Schulman, Sidor, Wu, and Zhokhov]{dhariwal2017openai}
P.~Dhariwal, C.~Hesse, O.~Klimov, A.~Nichol, M.~Plappert, A.~Radford, J.~Schulman, S.~Sidor, Y.~Wu, and P.~Zhokhov.
\newblock {OpenAI Baselines}.
\newblock \url{https://github.com/openai/baselines}, 2017.
\newblock URL \url{https://github.com/openai/baselines}.

\bibitem[Ecoffet et~al.(2021)Ecoffet, Huizinga, Lehman, Stanley, and Clune]{ecoffet2021first}
A.~Ecoffet, J.~Huizinga, J.~Lehman, K.~O. Stanley, and J.~Clune.
\newblock {First Return, Then Explore}.
\newblock \emph{Nature}, 590\penalty0 (7847):\penalty0 580--586, 2021.
\newblock \doi{10.1038/S41586-020-03157-9}.
\newblock URL \url{https://doi.org/10.1038/s41586-020-03157-9}.

\bibitem[Engstrom et~al.(2020)Engstrom, Ilyas, Santurkar, Tsipras, Janoos, Rudolph, and Madry]{engstrom2020implementation}
L.~Engstrom, A.~Ilyas, S.~Santurkar, D.~Tsipras, F.~Janoos, L.~Rudolph, and A.~Madry.
\newblock {Implementation Matters in Deep RL: A Case Study on PPO and TRPO}.
\newblock In \emph{8th International Conference on Learning Representations, {ICLR} 2020, Addis Ababa, Ethiopia, April 26-30, 2020}. OpenReview.net, 2020.
\newblock URL \url{https://openreview.net/forum?id=r1etN1rtPB}.

\bibitem[Espeholt et~al.(2018)Espeholt, Soyer, Munos, Simonyan, Mnih, Ward, Doron, Firoiu, Harley, Dunning, Legg, and Kavukcuoglu]{espeholt2018impala}
L.~Espeholt, H.~Soyer, R.~Munos, K.~Simonyan, V.~Mnih, T.~Ward, Y.~Doron, V.~Firoiu, T.~Harley, I.~Dunning, S.~Legg, and K.~Kavukcuoglu.
\newblock {IMPALA: Scalable Distributed Deep-RL with Importance Weighted Actor-Learner Architectures}.
\newblock In J.~G. Dy and A.~Krause, editors, \emph{Proceedings of the 35th International Conference on Machine Learning, {ICML} 2018, Stockholmsm{\"{a}}ssan, Stockholm, Sweden, July 10-15, 2018}, volume~80 of \emph{Proceedings of Machine Learning Research}, pages 1406--1415. {PMLR}, 2018.
\newblock URL \url{http://proceedings.mlr.press/v80/espeholt18a.html}.

\bibitem[Felten et~al.(2023)Felten, Alegre, Nowe, Bazzan, Talbi, Danoy, and da~Silva]{felten2023toolkit}
F.~Felten, L.~N. Alegre, A.~Nowe, A.~L.~C. Bazzan, E.~G. Talbi, G.~Danoy, and B.~C. da~Silva.
\newblock {A Toolkit for Reliable Benchmarking and Research in Multi-Objective Reinforcement Learning}.
\newblock In \emph{Proceedings of the Neural Information Processing Systems Track on Datasets and Benchmarks 3, NeurIPS Datasets and Benchmarks 2023}, 2023.
\newblock URL \url{https://openreview.net/forum?id=jfwRLudQyj}.

\bibitem[Fujimoto et~al.(2018)Fujimoto, van Hoof, and Meger]{fujimoto2018addressing}
S.~Fujimoto, H.~van Hoof, and D.~Meger.
\newblock {Addressing Function Approximation Error in Actor-Critic Methods}.
\newblock In J.~G. Dy and A.~Krause, editors, \emph{Proceedings of the 35th International Conference on Machine Learning, {ICML} 2018, Stockholmsm{\"{a}}ssan, Stockholm, Sweden, July 10-15, 2018}, volume~80 of \emph{Proceedings of Machine Learning Research}, pages 1582--1591. {PMLR}, 2018.
\newblock URL \url{http://proceedings.mlr.press/v80/fujimoto18a.html}.

\bibitem[Fujita et~al.(2021)Fujita, Nagarajan, Kataoka, and Ishikawa]{fujita2021chainerrl}
Y.~Fujita, P.~Nagarajan, T.~Kataoka, and T.~Ishikawa.
\newblock {ChainerRL: A Deep Reinforcement Learning Library}.
\newblock \emph{Journal of Machine Learning Research}, 22\penalty0 (77):\penalty0 1--14, 2021.
\newblock URL \url{http://jmlr.org/papers/v22/20-376.html}.

\bibitem[Gallouédec et~al.(2021)Gallouédec, Cazin, Dellandr{\'e}a, and Chen]{gallouédec2021panda}
Q.~Gallouédec, N.~Cazin, E.~Dellandr{\'e}a, and L.~Chen.
\newblock {panda-gym: Open-Source Goal-Conditioned Environments for Robotic Learning}.
\newblock \emph{4th Robot Learning Workshop: Self-Supervised and Lifelong Learning at NeurIPS}, 2021.

\bibitem[garage contributors(2019)]{contributors2019garage}
T.~garage contributors.
\newblock {Garage: A toolkit for reproducible reinforcement learning research}.
\newblock \url{https://github.com/rlworkgroup/garage}, 2019.

\bibitem[Guadarrama et~al.(2018)Guadarrama, Korattikara, Ramirez, Castro, Holly, Fishman, Wang, Gonina, Wu, Kokiopoulou, Sbaiz, Smith, Bartók, Berent, Harris, Vanhoucke, and Brevdo]{guadarrama2018tf}
S.~Guadarrama, A.~Korattikara, O.~Ramirez, P.~Castro, E.~Holly, S.~Fishman, K.~Wang, E.~Gonina, N.~Wu, E.~Kokiopoulou, L.~Sbaiz, J.~Smith, G.~Bartók, J.~Berent, C.~Harris, V.~Vanhoucke, and E.~Brevdo.
\newblock {TF-Agents: A library for Reinforcement Learning in TensorFlow}.
\newblock \url{https://github.com/tensorflow/agents}, 2018.
\newblock URL \url{https://github.com/tensorflow/agents}.

\bibitem[Haarnoja et~al.(2018)Haarnoja, Zhou, Abbeel, and Levine]{haarnoja2018soft}
T.~Haarnoja, A.~Zhou, P.~Abbeel, and S.~Levine.
\newblock {Soft Actor-Critic: Off-Policy Maximum Entropy Deep Reinforcement Learning with a Stochastic Actor}.
\newblock In J.~G. Dy and A.~Krause, editors, \emph{Proceedings of the 35th International Conference on Machine Learning, {ICML} 2018, Stockholmsm{\"{a}}ssan, Stockholm, Sweden, July 10-15, 2018}, volume~80 of \emph{Proceedings of Machine Learning Research}, pages 1856--1865. {PMLR}, 2018.
\newblock URL \url{http://proceedings.mlr.press/v80/haarnoja18b.html}.

\bibitem[Hafner et~al.(2023)Hafner, Pasukonis, Ba, and Lillicrap]{hafner2023mastering}
D.~Hafner, J.~Pasukonis, J.~Ba, and T.~Lillicrap.
\newblock {Mastering Diverse Domains through World Models}.
\newblock \emph{arXiv preprint arXiv:2301.04104}, 2023.

\bibitem[Henderson et~al.(2018)Henderson, Islam, Bachman, Pineau, Precup, and Meger]{henderson2018deep}
P.~Henderson, R.~Islam, P.~Bachman, J.~Pineau, D.~Precup, and D.~Meger.
\newblock {Deep Reinforcement Learning That Matters}.
\newblock In S.~A. McIlraith and K.~Q. Weinberger, editors, \emph{Proceedings of the Thirty-Second {AAAI} Conference on Artificial Intelligence, (AAAI-18), the 30th innovative Applications of Artificial Intelligence (IAAI-18), and the 8th {AAAI} Symposium on Educational Advances in Artificial Intelligence (EAAI-18), New Orleans, Louisiana, USA, February 2-7, 2018}, pages 3207--3214. {AAAI} Press, 2018.
\newblock \doi{10.1609/AAAI.V32I1.11694}.
\newblock URL \url{https://doi.org/10.1609/aaai.v32i1.11694}.

\bibitem[Hessel et~al.(2018)Hessel, Modayil, van Hasselt, Schaul, Ostrovski, Dabney, Horgan, Piot, Azar, and Silver]{hessel2018rainbow}
M.~Hessel, J.~Modayil, H.~van Hasselt, T.~Schaul, G.~Ostrovski, W.~Dabney, D.~Horgan, B.~Piot, M.~G. Azar, and D.~Silver.
\newblock {Rainbow: Combining Improvements in Deep Reinforcement Learning}.
\newblock In S.~A. McIlraith and K.~Q. Weinberger, editors, \emph{Proceedings of the Thirty-Second {AAAI} Conference on Artificial Intelligence, (AAAI-18), the 30th innovative Applications of Artificial Intelligence (IAAI-18), and the 8th {AAAI} Symposium on Educational Advances in Artificial Intelligence (EAAI-18), New Orleans, Louisiana, USA, February 2-7, 2018}, pages 3215--3222. {AAAI} Press, 2018.
\newblock \doi{10.1609/AAAI.V32I1.11796}.
\newblock URL \url{https://doi.org/10.1609/aaai.v32i1.11796}.

\bibitem[Hoffman et~al.(2020)Hoffman, Shahriari, Aslanides, Barth-Maron, Momchev, Sinopalnikov, Sta\'nczyk, Ramos, Raichuk, Vincent, Hussenot, Dadashi, Dulac-Arnold, Orsini, Jacq, Ferret, Vieillard, Ghasemipour, Girgin, Pietquin, Behbahani, Norman, Abdolmaleki, Cassirer, Yang, Baumli, Henderson, Friesen, Haroun, Novikov, Colmenarejo, Cabi, Gulcehre, Paine, Srinivasan, Cowie, Wang, Piot, and de~Freitas]{hoffman2020acme}
M.~W. Hoffman, B.~Shahriari, J.~Aslanides, G.~Barth-Maron, N.~Momchev, D.~Sinopalnikov, P.~Sta\'nczyk, S.~Ramos, A.~Raichuk, D.~Vincent, L.~Hussenot, R.~Dadashi, G.~Dulac-Arnold, M.~Orsini, A.~Jacq, J.~Ferret, N.~Vieillard, S.~K.~S. Ghasemipour, S.~Girgin, O.~Pietquin, F.~Behbahani, T.~Norman, A.~Abdolmaleki, A.~Cassirer, F.~Yang, K.~Baumli, S.~Henderson, A.~Friesen, R.~Haroun, A.~Novikov, S.~G. Colmenarejo, S.~Cabi, C.~Gulcehre, T.~L. Paine, S.~Srinivasan, A.~Cowie, Z.~Wang, B.~Piot, and N.~de~Freitas.
\newblock {Acme: A Research Framework for Distributed Reinforcement Learning}.
\newblock \emph{arXiv preprint arXiv:2006.00979}, 2020.

\bibitem[Huang et~al.(2022{\natexlab{a}})Huang, Dossa, Raffin, Kanervisto, and Wang]{huang202237}
S.~Huang, R.~F.~J. Dossa, A.~Raffin, A.~Kanervisto, and W.~Wang.
\newblock {The 37 Implementation Details of Proximal Policy Optimization}.
\newblock In \emph{ICLR Blog Track}, 2022{\natexlab{a}}.
\newblock URL \url{https://iclr-blog-track.github.io/2022/03/25/ppo-implementation-details/}.
\newblock https://iclr-blog-track.github.io/2022/03/25/ppo-implementation-details/.

\bibitem[Huang et~al.(2022{\natexlab{b}})Huang, Dossa, Ye, Braga, Chakraborty, Mehta, and Araújo]{huang2022cleanrl}
S.~Huang, R.~F.~J. Dossa, C.~Ye, J.~Braga, D.~Chakraborty, K.~Mehta, and J.~G. Araújo.
\newblock {CleanRL: High-quality Single-file Implementations of Deep Reinforcement Learning Algorithms}.
\newblock \emph{Journal of Machine Learning Research}, 23\penalty0 (274):\penalty0 1--18, 2022{\natexlab{b}}.
\newblock URL \url{http://jmlr.org/papers/v23/21-1342.html}.

\bibitem[Huang et~al.(2023)Huang, Weng, Charakorn, Lin, Xu, and Ontañón]{huang2023cleanba}
S.~Huang, J.~Weng, R.~Charakorn, M.~Lin, Z.~Xu, and S.~Ontañón.
\newblock {Cleanba: A Reproducible and Efficient Distributed Reinforcement Learning Platform}, 2023.

\bibitem[Janner et~al.(2019)Janner, Fu, Zhang, and Levine]{janner2019when}
M.~Janner, J.~Fu, M.~Zhang, and S.~Levine.
\newblock {When to Trust Your Model: Model-Based Policy Optimization}.
\newblock In H.~M. Wallach, H.~Larochelle, A.~Beygelzimer, F.~d'Alch{\'{e}}{-}Buc, E.~B. Fox, and R.~Garnett, editors, \emph{Advances in Neural Information Processing Systems 32: Annual Conference on Neural Information Processing Systems 2019, NeurIPS 2019, December 8-14, 2019, Vancouver, BC, Canada}, pages 12498--12509, 2019.
\newblock URL \url{https://proceedings.neurips.cc/paper/2019/hash/5faf461eff3099671ad63c6f3f094f7f-Abstract.html}.

\bibitem[Kostrikov(2021)]{kostrikov2021jaxrl}
I.~Kostrikov.
\newblock {JAXRL: Implementations of Reinforcement Learning algorithms in JAX}.
\newblock \url{https://github.com/ikostrikov/jaxrl}, Oct 2021.
\newblock URL \url{https://github.com/ikostrikov/jaxrl}.

\bibitem[Kuhnle et~al.(2017)Kuhnle, Schaarschmidt, and Fricke]{kuhnle2017tensorforce}
A.~Kuhnle, M.~Schaarschmidt, and K.~Fricke.
\newblock {Tensorforce: a TensorFlow library for applied reinforcement learning}.
\newblock \url{https://github.com/tensorforce/tensorforce}, 2017.
\newblock URL \url{https://github.com/tensorforce/tensorforce}.

\bibitem[K{\"{u}}ttler et~al.(2019)K{\"{u}}ttler, Nardelli, Lavril, Selvatici, Sivakumar, Rockt{\"{a}}schel, and Grefenstette]{k_uttler2019torchbeast}
H.~K{\"{u}}ttler, N.~Nardelli, T.~Lavril, M.~Selvatici, V.~Sivakumar, T.~Rockt{\"{a}}schel, and E.~Grefenstette.
\newblock {TorchBeast: A PyTorch Platform for Distributed RL}.
\newblock \emph{arXiv preprint arXiv:1910.03552}, 2019.

\bibitem[Lee et~al.(2022)Lee, Nachum, Yang, Lee, Freeman, Guadarrama, Fischer, Xu, Jang, Michalewski, and Mordatch]{lee2022multi}
K.~Lee, O.~Nachum, M.~Yang, L.~Lee, D.~Freeman, S.~Guadarrama, I.~Fischer, W.~Xu, E.~Jang, H.~Michalewski, and I.~Mordatch.
\newblock {Multi-Game Decision Transformers}.
\newblock In S.~Koyejo, S.~Mohamed, A.~Agarwal, D.~Belgrave, K.~Cho, and A.~Oh, editors, \emph{Advances in Neural Information Processing Systems 35: Annual Conference on Neural Information Processing Systems 2022, NeurIPS 2022, New Orleans, LA, USA, November 28 - December 9, 2022}, 2022.
\newblock URL \url{http://papers.nips.cc/paper\_files/paper/2022/hash/b2cac94f82928a85055987d9fd44753f-Abstract-Conference.html}.

\bibitem[Leurent(2018)]{leurent2018environment}
E.~Leurent.
\newblock {An Environment for Autonomous Driving Decision-Making}.
\newblock \url{https://github.com/eleurent/highway-env}, 2018.
\newblock URL \url{https://github.com/eleurent/highway-env}.

\bibitem[Liang et~al.(2018)Liang, Liaw, Nishihara, Moritz, Fox, Goldberg, Gonzalez, Jordan, and Stoica]{liang2018rllib}
E.~Liang, R.~Liaw, R.~Nishihara, P.~Moritz, R.~Fox, K.~Goldberg, J.~Gonzalez, M.~I. Jordan, and I.~Stoica.
\newblock {RLlib: Abstractions for Distributed Reinforcement Learning}.
\newblock In J.~G. Dy and A.~Krause, editors, \emph{Proceedings of the 35th International Conference on Machine Learning, {ICML} 2018, Stockholmsm{\"{a}}ssan, Stockholm, Sweden, July 10-15, 2018}, volume~80 of \emph{Proceedings of Machine Learning Research}, pages 3059--3068. {PMLR}, 2018.
\newblock URL \url{http://proceedings.mlr.press/v80/liang18b.html}.

\bibitem[Lynnerup et~al.(2019)Lynnerup, Nolling, Hasle, and Hallam]{lynnerup2019survey}
N.~A. Lynnerup, L.~Nolling, R.~Hasle, and J.~Hallam.
\newblock {A Survey on Reproducibility by Evaluating Deep Reinforcement Learning Algorithms on Real-World Robots}.
\newblock In L.~P. Kaelbling, D.~Kragic, and K.~Sugiura, editors, \emph{3rd Annual Conference on Robot Learning, CoRL 2019, Osaka, Japan, October 30 - November 1, 2019, Proceedings}, volume 100 of \emph{Proceedings of Machine Learning Research}, pages 466--489. {PMLR}, 2019.
\newblock URL \url{http://proceedings.mlr.press/v100/lynnerup20a.html}.

\bibitem[Machado et~al.(2018)Machado, Bellemare, Talvitie, Veness, Hausknecht, and Bowling]{machado2018revisiting}
M.~C. Machado, M.~G. Bellemare, E.~Talvitie, J.~Veness, M.~J. Hausknecht, and M.~Bowling.
\newblock {Revisiting the Arcade Learning Environment: Evaluation Protocols and Open Problems for General Agents}.
\newblock \emph{Journal of Artificial Intelligence Research}, 61:\penalty0 523--562, 2018.
\newblock \doi{10.1613/JAIR.5699}.
\newblock URL \url{https://doi.org/10.1613/jair.5699}.

\bibitem[Makoviichuk and Makoviychuk(2021)]{makoviichuk2021rl}
D.~Makoviichuk and V.~Makoviychuk.
\newblock {rl-games: A High-performance Framework for Reinforcement Learning}.
\newblock \url{https://github.com/Denys88/rl_games}, May 2021.
\newblock URL \url{https://github.com/Denys88/rl_games}.

\bibitem[Mella et~al.(2022)Mella, Hambro, Rothermel, and K{\"{u}}ttler]{mella2022moolib}
V.~Mella, E.~Hambro, D.~Rothermel, and H.~K{\"{u}}ttler.
\newblock {moolib: A Platform for Distributed RL}.
\newblock \emph{GitHub repository}, 2022.
\newblock URL \url{https://github.com/facebookresearch/moolib}.

\bibitem[Mnih et~al.(2013)Mnih, Kavukcuoglu, Silver, Graves, Antonoglou, Wierstra, and Riedmiller]{mnih2013playing}
V.~Mnih, K.~Kavukcuoglu, D.~Silver, A.~Graves, I.~Antonoglou, D.~Wierstra, and M.~A. Riedmiller.
\newblock {Playing Atari with Deep Reinforcement Learning}.
\newblock \emph{arXiv preprint arXiv:1312.5602}, 2013.

\bibitem[Mnih et~al.(2016)Mnih, Badia, Mirza, Graves, Lillicrap, Harley, Silver, and Kavukcuoglu]{mnih2016asynchronous}
V.~Mnih, A.~P. Badia, M.~Mirza, A.~Graves, T.~P. Lillicrap, T.~Harley, D.~Silver, and K.~Kavukcuoglu.
\newblock {Asynchronous Methods for Deep Reinforcement Learning}.
\newblock In M.~Balcan and K.~Q. Weinberger, editors, \emph{Proceedings of the 33nd International Conference on Machine Learning, {ICML} 2016, New York City, NY, USA, June 19-24, 2016}, volume~48 of \emph{{JMLR} Workshop and Conference Proceedings}, pages 1928--1937. JMLR.org, 2016.
\newblock URL \url{http://proceedings.mlr.press/v48/mniha16.html}.

\bibitem[Patterson et~al.(2023)Patterson, Neumann, White, and White]{patterson2023empirical}
A.~Patterson, S.~Neumann, M.~White, and A.~White.
\newblock {Empirical Design in Reinforcement Learning}.
\newblock \emph{arXiv preprint arXiv:2304.01315}, 2023.

\bibitem[Petrenko et~al.(2020)Petrenko, Huang, Kumar, Sukhatme, and Koltun]{petrenko2020sample}
A.~Petrenko, Z.~Huang, T.~Kumar, G.~S. Sukhatme, and V.~Koltun.
\newblock {Sample Factory: Egocentric 3D Control from Pixels at 100000 FPS with Asynchronous Reinforcement Learning}.
\newblock In \emph{Proceedings of the 37th International Conference on Machine Learning, {ICML} 2020, 13-18 July 2020, Virtual Event}, volume 119 of \emph{Proceedings of Machine Learning Research}, pages 7652--7662. {PMLR}, 2020.
\newblock URL \url{http://proceedings.mlr.press/v119/petrenko20a.html}.

\bibitem[Pineau et~al.(2021)Pineau, Vincent{-}Lamarre, Sinha, Larivi{\`{e}}re, Beygelzimer, d'Alch{\'{e}}{-}Buc, Fox, and Larochelle]{pineau2021improving}
J.~Pineau, P.~Vincent{-}Lamarre, K.~Sinha, V.~Larivi{\`{e}}re, A.~Beygelzimer, F.~d'Alch{\'{e}}{-}Buc, E.~B. Fox, and H.~Larochelle.
\newblock {Improving Reproducibility in Machine Learning Research (A Report from the NeurIPS 2019 Reproducibility Program)}.
\newblock \emph{Journal of Machine Learning Research}, 22:\penalty0 164:1--164:20, 2021.
\newblock URL \url{http://jmlr.org/papers/v22/20-303.html}.

\bibitem[Plappert et~al.(2018)Plappert, Andrychowicz, Ray, McGrew, Baker, Powell, Schneider, Tobin, Chociej, Welinder, Kumar, and Zaremba]{plappert2018multi}
M.~Plappert, M.~Andrychowicz, A.~Ray, B.~McGrew, B.~Baker, G.~Powell, J.~Schneider, J.~Tobin, M.~Chociej, P.~Welinder, V.~Kumar, and W.~Zaremba.
\newblock {Multi-Goal Reinforcement Learning: Challenging Robotics Environments and Request for Research}.
\newblock \emph{arXiv preprint arXiv:1802.09464}, 2018.

\bibitem[Raffin(2020)]{raffin2020rl}
A.~Raffin.
\newblock {RL Baselines3 Zoo}.
\newblock \url{https://github.com/DLR-RM/rl-baselines3-zoo}, 2020.

\bibitem[Raffin et~al.(2021)Raffin, Hill, Gleave, Kanervisto, Ernestus, and Dormann]{raffin2021stable}
A.~Raffin, A.~Hill, A.~Gleave, A.~Kanervisto, M.~Ernestus, and N.~Dormann.
\newblock {Stable-Baselines3: Reliable Reinforcement Learning Implementations}.
\newblock \emph{Journal of Machine Learning Research}, 22\penalty0 (268):\penalty0 1--8, 2021.

\bibitem[Reed et~al.(2022)Reed, Zolna, Parisotto, Colmenarejo, Novikov, Barth{-}Maron, Gimenez, Sulsky, Kay, Springenberg, Eccles, Bruce, Razavi, Edwards, Heess, Chen, Hadsell, Vinyals, Bordbar, and de~Freitas]{reed2022generalist}
S.~E. Reed, K.~Zolna, E.~Parisotto, S.~G. Colmenarejo, A.~Novikov, G.~Barth{-}Maron, M.~Gimenez, Y.~Sulsky, J.~Kay, J.~T. Springenberg, T.~Eccles, J.~Bruce, A.~Razavi, A.~Edwards, N.~Heess, Y.~Chen, R.~Hadsell, O.~Vinyals, M.~Bordbar, and N.~de~Freitas.
\newblock {A Generalist Agent}.
\newblock \emph{Transactions on Machine Learning Research}, 2022, 2022.
\newblock URL \url{https://openreview.net/forum?id=1ikK0kHjvj}.

\bibitem[Schulman et~al.(2015)Schulman, Levine, Abbeel, Jordan, and Moritz]{schulman2015trust}
J.~Schulman, S.~Levine, P.~Abbeel, M.~I. Jordan, and P.~Moritz.
\newblock {Trust Region Policy Optimization}.
\newblock In F.~R. Bach and D.~M. Blei, editors, \emph{Proceedings of the 32nd International Conference on Machine Learning, {ICML} 2015, Lille, France, 6-11 July 2015}, volume~37 of \emph{{JMLR} Workshop and Conference Proceedings}, pages 1889--1897. JMLR.org, 2015.
\newblock URL \url{http://proceedings.mlr.press/v37/schulman15.html}.

\bibitem[Schulman et~al.(2016)Schulman, Moritz, Levine, Jordan, and Abbeel]{schulman2016high}
J.~Schulman, P.~Moritz, S.~Levine, M.~I. Jordan, and P.~Abbeel.
\newblock {High-Dimensional Continuous Control Using Generalized Advantage Estimation}.
\newblock In Y.~Bengio and Y.~LeCun, editors, \emph{4th International Conference on Learning Representations, {ICLR} 2016, San Juan, Puerto Rico, May 2-4, 2016, Conference Track Proceedings}, 2016.
\newblock URL \url{http://arxiv.org/abs/1506.02438}.

\bibitem[Schulman et~al.(2017)Schulman, Wolski, Dhariwal, Radford, and Klimov]{schulman2017proximal}
J.~Schulman, F.~Wolski, P.~Dhariwal, A.~Radford, and O.~Klimov.
\newblock {Proximal Policy Optimization Algorithms}.
\newblock \emph{arXiv preprint arXiv:1707.06347}, 2017.

\bibitem[Todorov et~al.(2012)Todorov, Erez, and Tassa]{todorov2012mujoco}
E.~Todorov, T.~Erez, and Y.~Tassa.
\newblock {MuJoCo: A Physics Engine for Model-Based Control}.
\newblock In \emph{2012 {IEEE/RSJ} International Conference on Intelligent Robots and Systems, {IROS} 2012, Vilamoura, Algarve, Portugal, October 7-12, 2012}, pages 5026--5033. {IEEE}, 2012.

\bibitem[Toromanoff et~al.(2019)Toromanoff, Wirbel, and Moutarde]{toromanoff2019is}
M.~Toromanoff, {\'{E}}.~Wirbel, and F.~Moutarde.
\newblock {Is Deep Reinforcement Learning Really Superhuman on Atari?}
\newblock \emph{arXiv preprint arXiv:1908.04683}, 2019.

\bibitem[Towers et~al.(2023)Towers, Terry, Kwiatkowski, Balis, Cola, Deleu, Goulão, Kallinteris, KG, Krimmel, Perez-Vicente, Pierré, Schulhoff, Tai, Shen, and Younis]{towers2023gymnasium}
M.~Towers, J.~K. Terry, A.~Kwiatkowski, J.~U. Balis, G.~d. Cola, T.~Deleu, M.~Goulão, A.~Kallinteris, A.~KG, M.~Krimmel, R.~Perez-Vicente, A.~Pierré, S.~Schulhoff, J.~J. Tai, A.~T.~J. Shen, and O.~G. Younis.
\newblock {Gymnasium}, Mar. 2023.
\newblock URL \url{https://zenodo.org/record/8127025}.

\bibitem[Weng et~al.(2022{\natexlab{a}})Weng, Chen, Yan, You, Duburcq, Zhang, Su, Su, and Zhu]{weng2022tianshou}
J.~Weng, H.~Chen, D.~Yan, K.~You, A.~Duburcq, M.~Zhang, Y.~Su, H.~Su, and J.~Zhu.
\newblock {Tianshou: A Highly Modularized Deep Reinforcement Learning Library}.
\newblock \emph{Journal of Machine Learning Research}, 23\penalty0 (267):\penalty0 1--6, 2022{\natexlab{a}}.
\newblock URL \url{http://jmlr.org/papers/v23/21-1127.html}.

\bibitem[Weng et~al.(2022{\natexlab{b}})Weng, Lin, Huang, Liu, Makoviichuk, Makoviychuk, Liu, Song, Luo, Jiang, Xu, and Yan]{weng2022envpool}
J.~Weng, M.~Lin, S.~Huang, B.~Liu, D.~Makoviichuk, V.~Makoviychuk, Z.~Liu, Y.~Song, T.~Luo, Y.~Jiang, Z.~Xu, and S.~Yan.
\newblock {EnvPool: A Highly Parallel Reinforcement Learning Environment Execution Engine}.
\newblock In \emph{Proceedings of the Neural Information Processing Systems Track on Datasets and Benchmarks 2, NeurIPS Datasets and Benchmarks 2022}, 2022{\natexlab{b}}.
\newblock URL \url{http://papers.nips.cc/paper\_files/paper/2022/hash/8caaf08e49ddbad6694fae067442ee21-Abstract-Datasets\_and\_Benchmarks.html}.

\bibitem[Zhao(2022)]{zhao2022abcdrl}
Y.~Zhao.
\newblock {abcdRL: Modular Single-file Reinforcement Learning Algorithms Library}.
\newblock \url{https://github.com/sdpkjc/abcdrl}, Dec. 2022.
\newblock URL \url{https://github.com/sdpkjc/abcdrl}.

\end{thebibliography}

\appendix
\clearpage
\section{Plotting Results Guidelines}

\subsection{Using the CLI}
\label{sec:appendix_cli}

This section gives notable additional examples of usage of the provided CLI. A more comprehensive set of examples and manual is available in the README page of the project.

\subsubsection{Plotting episodic return from various libraries}
\begin{figure}[ht]
    \centering
    \includegraphics[width=0.95\textwidth]{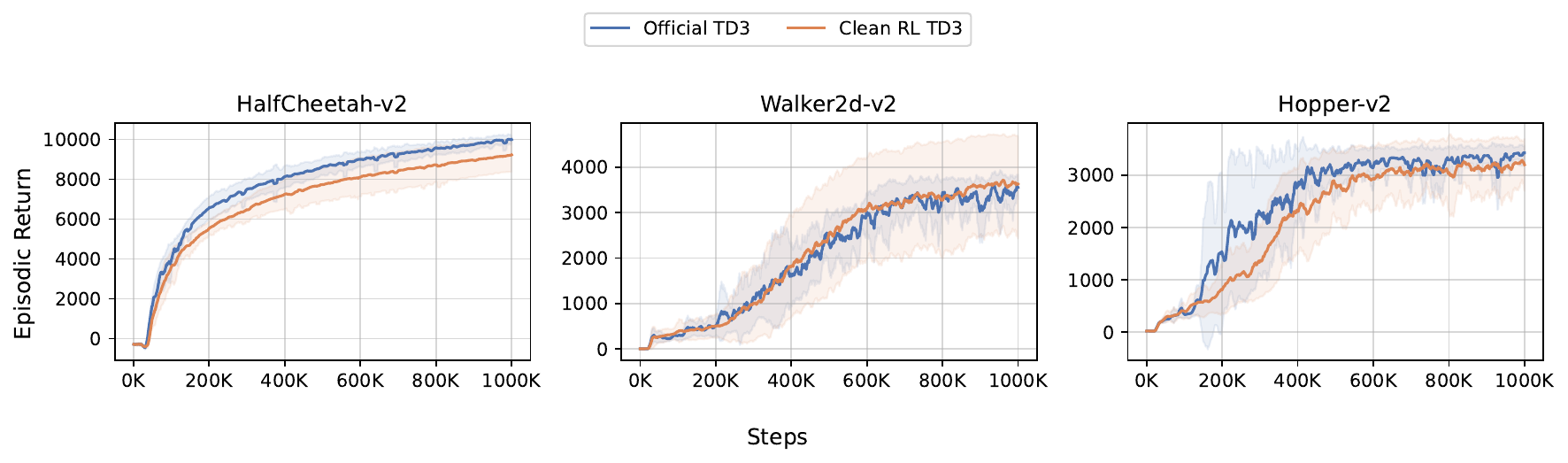}
    \caption{Comparing CleanRL's TD3 against the original TD3 implementation (sample efficiency).}
    \label{fig:cleanrl_vs_td3}
\end{figure}

\begin{figure}[ht]
    \centering
    \includegraphics[width=0.95\textwidth]{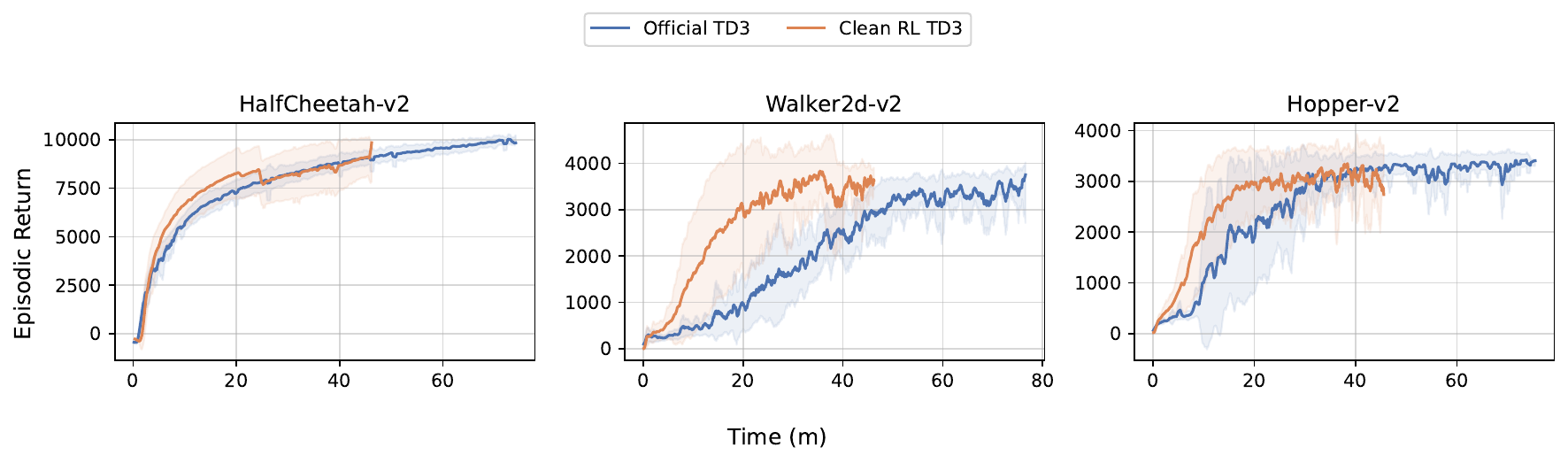}
    \caption{Comparing CleanRL's TD3 against the original TD3 implementation (time).}
    \label{fig:cleanrl_vs_td3 time}
\end{figure}

First, we showcase the most basic usage of the CLI, that is comparing two different implementations of the same algorithm based on learning curve of episodic return. For example, Figure~\ref{fig:cleanrl_vs_td3} and Figure~\ref{fig:cleanrl_vs_td3 time} compare CleanRL's TD3 implementation against the original TD3, both in terms of sample efficiency and time. The command used to generate this plot is listed below.

\lstset{
  columns=fullflexible,
  breaklines=true,
  moredelim=**[is][\color{red}]{@}{@},
}
\begin{lstlisting}
python -m openrlbenchmark.rlops \
    --filters '?we=openrlbenchmark&@wpn=sfujim-TD3@&ceik=env&cen=policy&@metric=charts/episodic_return@' '@TD3?cl=Official TD3@' \
    --filters '?we=openrlbenchmark&@wpn=cleanrl@&ceik=env_id&cen=exp_name&@metric=charts/episodic_return@' '@td3_continuous_action_jax?cl=Clean RL TD3@' \
    --env-ids HalfCheetah-v2 Walker2d-v2 Hopper-v2 \
    --pc.ncols 3 \
    --pc.ncols-legend 2 \
    --output-filename static/td3_vs_cleanrl \
    --scan-history
\end{lstlisting}

In the above command, \verb|wpn| denotes the project name, typically the learning library name. This allows to fetch results of implementations from different projects. Moreover, it is possible to specify which metric to compare, in this case \verb|charts/episodic_return|. Also, the CLI provides the possibility to select a given algorithm and apply a different name in the plot, e.g. we rename \verb|TD3| to \textit{Official TD3} and \verb|td3_continuous_action_jax| to \textit{Clean RL TD3}. Finally, we can also select a set of environments through the \verb|--end-ids| option. 

\subsubsection{RLiable integration}
\begin{figure}[ht]
    \centering
    \includegraphics[width=0.95\textwidth]{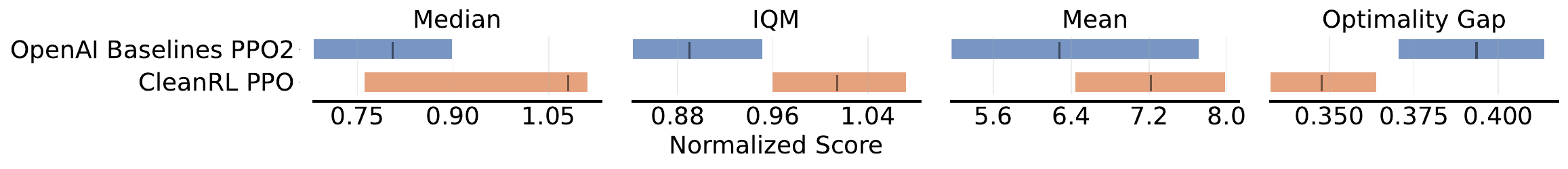}
    \caption{Clean RL PPO vs. OpenAI Baselines PPO, normalized score (RLiable).}
    \label{fig:cleanrl_ppo_vs_openai_aggregate}
\end{figure}

\begin{figure}[ht]
    \centering
    \includegraphics[width=0.95\textwidth]{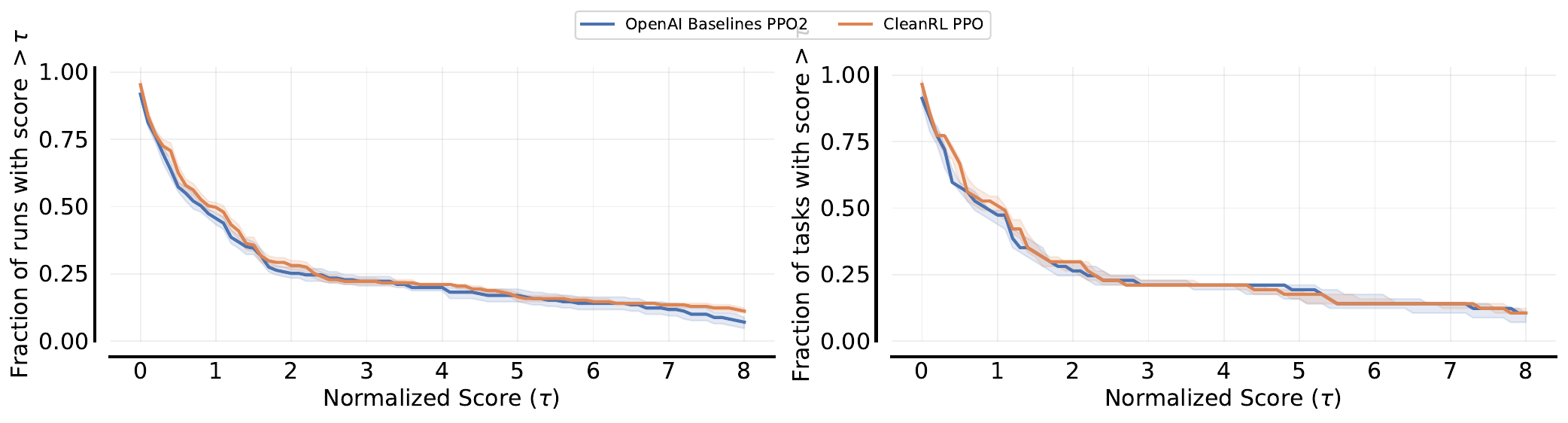}
    \caption{Clean RL PPO vs. OpenAI Baselines PPO, performance profile (RLiable).}
    \label{fig:cleanrl_ppo_vs_openai_profile}
\end{figure}

\begin{figure}[ht]
    \centering
    \includegraphics[width=0.95\textwidth]{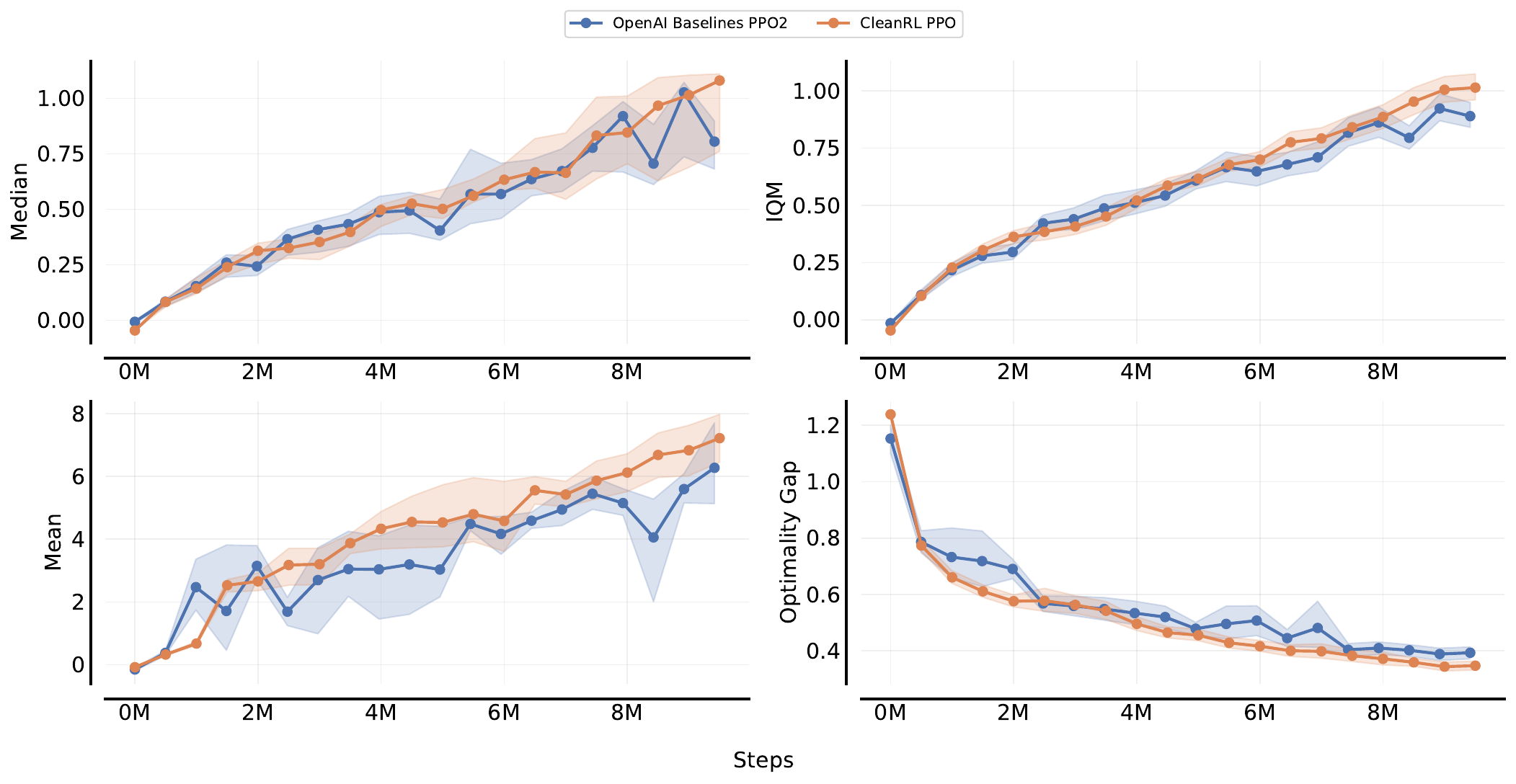}
    \caption{Clean RL PPO vs. OpenAI Baselines PPO, sample efficiency (RLiable).}
    \label{fig:cleanrl_ppo_vs_openai_sample}
\end{figure}

\begin{figure}[ht]
    \centering
    \includegraphics[width=0.95\textwidth]{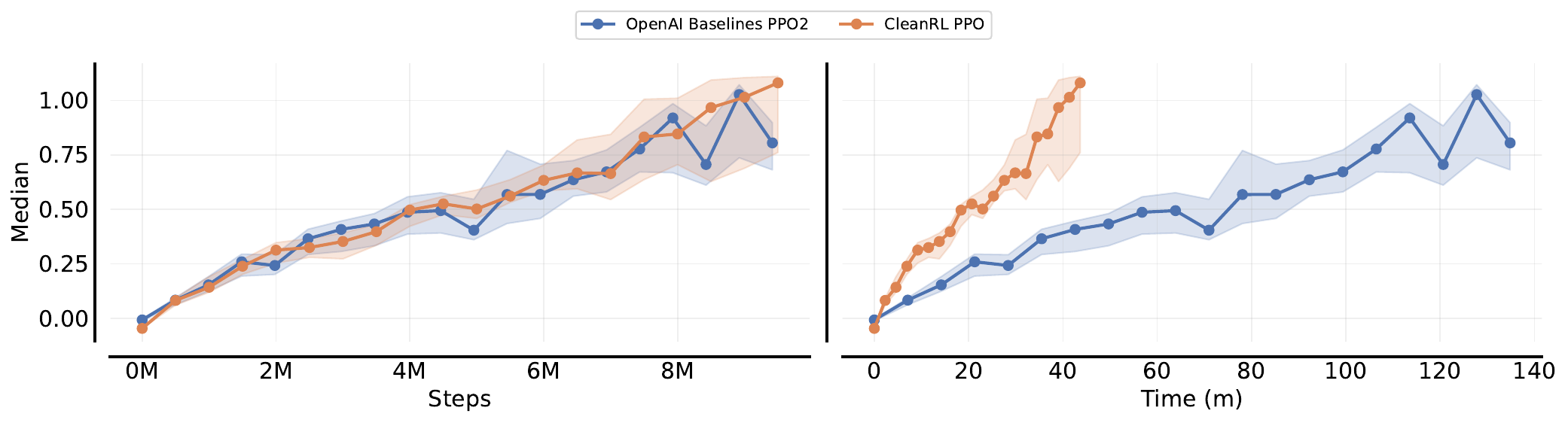}
    \caption{Clean RL PPO vs. OpenAI Baselines PPO, walltime efficiency (RLiable).}
    \label{fig:cleanrl_ppo_vs_openai_walltime}
\end{figure}

Open RL Benchmark also integrates with RLiable~\cite{agarwal2021deep}. To enable such plot, the option \verb|--rliable| can be toggled, then additional parameters are available under \verb|--rc|. Figures~\ref{fig:cleanrl_ppo_vs_openai_aggregate},~\ref{fig:cleanrl_ppo_vs_openai_profile},~\ref{fig:cleanrl_ppo_vs_openai_sample},~\ref{fig:cleanrl_ppo_vs_openai_walltime} showcase the resulting plots of the following command:

\begin{lstlisting}
python -m openrlbenchmark.rlops \
    --filters '?we=openrlbenchmark&wpn=baselines&ceik=env&cen=exp_name&metric=charts/episodic_return' 'baselines-ppo2-cnn?cl=OpenAI Baselines PPO2' \
    --filters '?we=openrlbenchmark&wpn=envpool-atari&ceik=env_id&cen=exp_name&metric=charts/avg_episodic_return' 'ppo_atari_envpool_xla_jax_truncation?cl=CleanRL PPO' \
    --env-ids AlienNoFrameskip-v4 AmidarNoFrameskip-v4 AssaultNoFrameskip-v4 AsterixNoFrameskip-v4 AsteroidsNoFrameskip-v4 AtlantisNoFrameskip-v4 BankHeistNoFrameskip-v4 BattleZoneNoFrameskip-v4 BeamRiderNoFrameskip-v4 BerzerkNoFrameskip-v4 BowlingNoFrameskip-v4 BoxingNoFrameskip-v4 BreakoutNoFrameskip-v4 CentipedeNoFrameskip-v4 ChopperCommandNoFrameskip-v4 CrazyClimberNoFrameskip-v4 DefenderNoFrameskip-v4 DemonAttackNoFrameskip-v4 DoubleDunkNoFrameskip-v4 EnduroNoFrameskip-v4 FishingDerbyNoFrameskip-v4 FreewayNoFrameskip-v4 FrostbiteNoFrameskip-v4 GopherNoFrameskip-v4 GravitarNoFrameskip-v4 HeroNoFrameskip-v4 IceHockeyNoFrameskip-v4 PrivateEyeNoFrameskip-v4 QbertNoFrameskip-v4 RiverraidNoFrameskip-v4 RoadRunnerNoFrameskip-v4 RobotankNoFrameskip-v4 SeaquestNoFrameskip-v4 SkiingNoFrameskip-v4 SolarisNoFrameskip-v4 SpaceInvadersNoFrameskip-v4 StarGunnerNoFrameskip-v4 SurroundNoFrameskip-v4 TennisNoFrameskip-v4 TimePilotNoFrameskip-v4 TutankhamNoFrameskip-v4 UpNDownNoFrameskip-v4 VentureNoFrameskip-v4 VideoPinballNoFrameskip-v4 WizardOfWorNoFrameskip-v4 YarsRevengeNoFrameskip-v4 ZaxxonNoFrameskip-v4 JamesbondNoFrameskip-v4 KangarooNoFrameskip-v4 KrullNoFrameskip-v4 KungFuMasterNoFrameskip-v4 MontezumaRevengeNoFrameskip-v4 MsPacmanNoFrameskip-v4 NameThisGameNoFrameskip-v4 PhoenixNoFrameskip-v4 PitfallNoFrameskip-v4 PongNoFrameskip-v4 \
    --env-ids Alien-v5 Amidar-v5 Assault-v5 Asterix-v5 Asteroids-v5 Atlantis-v5 BankHeist-v5 BattleZone-v5 BeamRider-v5 Berzerk-v5 Bowling-v5 Boxing-v5 Breakout-v5 Centipede-v5 ChopperCommand-v5 CrazyClimber-v5 Defender-v5 DemonAttack-v5 DoubleDunk-v5 Enduro-v5 FishingDerby-v5 Freeway-v5 Frostbite-v5 Gopher-v5 Gravitar-v5 Hero-v5 IceHockey-v5 PrivateEye-v5 Qbert-v5 Riverraid-v5 RoadRunner-v5 Robotank-v5 Seaquest-v5 Skiing-v5 Solaris-v5 SpaceInvaders-v5 StarGunner-v5 Surround-v5 Tennis-v5 TimePilot-v5 Tutankham-v5 UpNDown-v5 Venture-v5 VideoPinball-v5 WizardOfWor-v5 YarsRevenge-v5 Zaxxon-v5 Jamesbond-v5 Kangaroo-v5 Krull-v5 KungFuMaster-v5 MontezumaRevenge-v5 MsPacman-v5 NameThisGame-v5 Phoenix-v5 Pitfall-v5 Pong-v5 \
    --no-check-empty-runs \
    --pc.ncols 5 \
    --pc.ncols-legend 2 \
    @--rliable \
    --rc.score_normalization_method atari \
    --rc.normalized_score_threshold 8.0 \
    --rc.sample_efficiency_plots \
    --rc.sample_efficiency_and_walltime_efficiency_method Median \
    --rc.performance_profile_plots  \
    --rc.aggregate_metrics_plots  \
    --rc.sample_efficiency_num_bootstrap_reps 50000 \
    --rc.performance_profile_num_bootstrap_reps 2000 \
    --rc.interval_estimates_num_bootstrap_reps 2000@ \
    --output-filename static/cleanrl_vs_baselines_atari \
    --scan-history
\end{lstlisting}

\clearpage
\subsubsection{Multi-metrics}
\begin{figure}[ht]
    \centering
\includegraphics[width=0.95\textwidth]{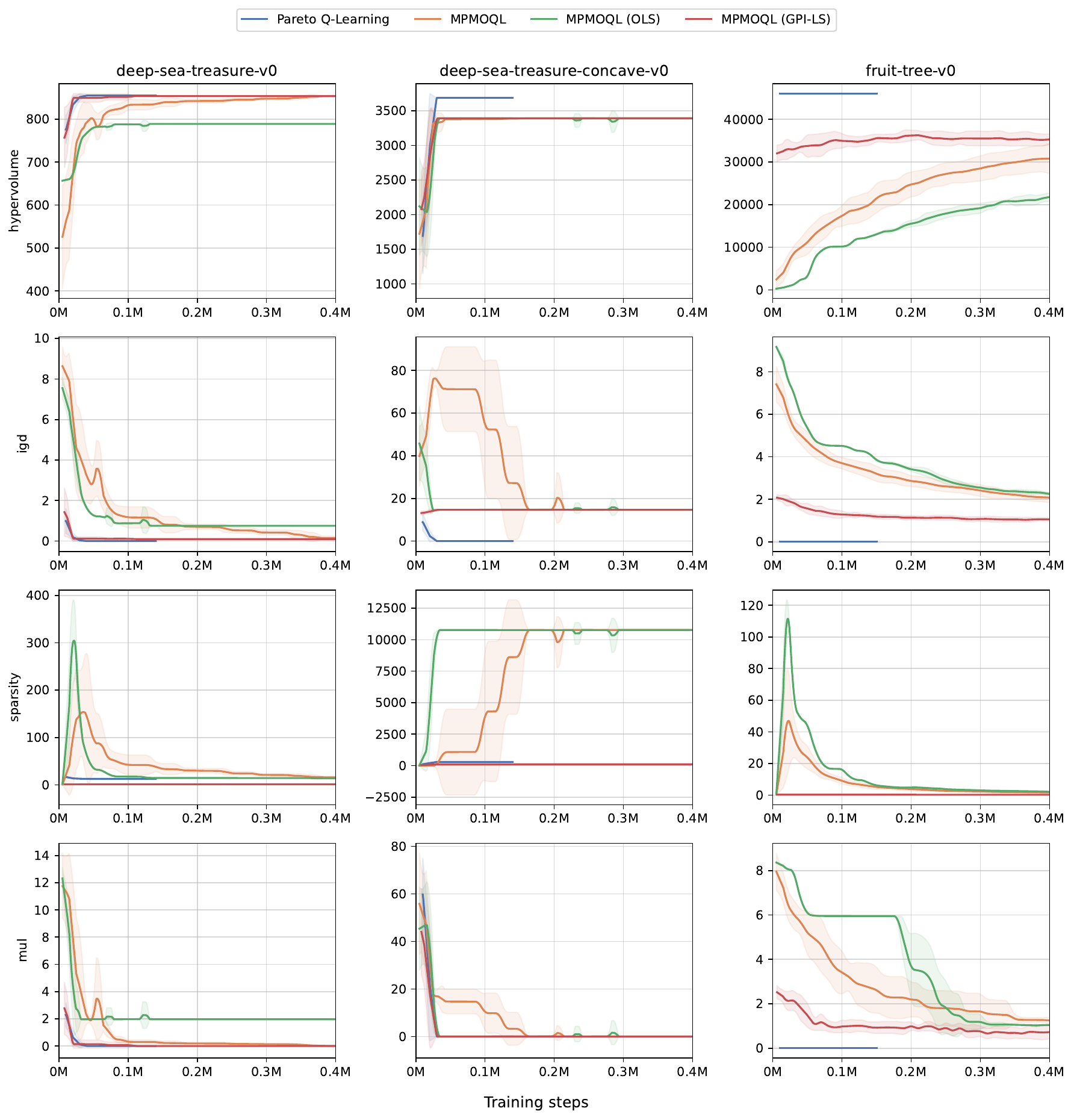}
    \caption{Plotting different metrics for different environments.}
    \label{fig:multi_metrics}
\end{figure}

Sometimes, such as in multi-objective RL (MORL), it is useful to report multiple metrics in the paper. Hence, the CLI includes an option to plot multiple metrics. Below is an example of CLI and resulting plots (Figure~\ref{fig:multi_metrics}) for multiple MORL algorithms on different environments.

\begin{lstlisting}
python -m openrlbenchmark@.rlops_multi_metrics@ \
  --filters '?we=openrlbenchmark&wpn=MORL-Baselines&ceik=env_id&cen=algo@&metrics=eval/hypervolume&metrics=eval/igd&metrics=eval/sparsity&metrics=eval/mul@' \
  'Pareto Q-Learning?cl=Pareto Q-Learning' \
  'MultiPolicy MO Q-Learning?cl=MPMOQL' \
  'MultiPolicy MO Q-Learning (OLS)?cl=MPMOQL (OLS)' \
  'MultiPolicy MO Q-Learning (GPI-LS)?cl=MPMOQL (GPI-LS)' \
  --env-ids deep-sea-treasure-v0 deep-sea-treasure-concave-v0 fruit-tree-v0 \
  --pc.ncols 3 \
  --pc.ncols-legend 4 \
  --pc.xlabel 'Training steps' \
  --pc.ylabel '' \
  --pc.max_steps 400000 \
  --output-filename morl/morl_deterministic_envs \
  --scan-history
\end{lstlisting}

\clearpage
\subsection{Using a custom script}
\label{sec:appendix_custom_script}
\begin{figure}[ht]
    \centering
    \includegraphics[width=0.5\textwidth]{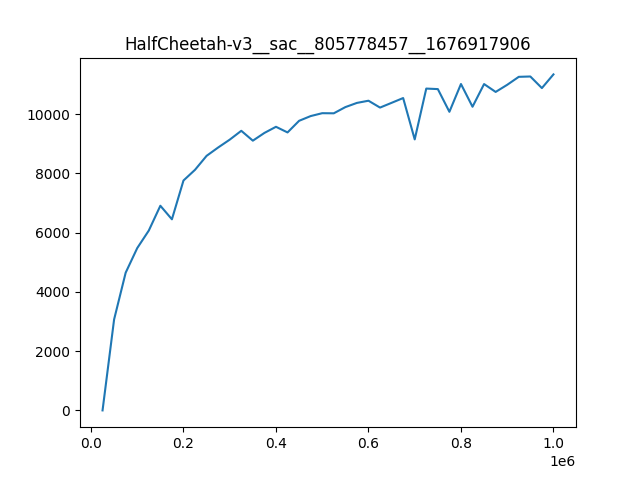}
    \caption{Example of a plot created with a custom script, by importing data directly from Open RL Benchmark using the WandB API.}
    \label{fig:custom_plot}
\end{figure}

Our CLI proves highly beneficial for generating standard RL plots, as demonstrated above. Nevertheless, in certain specialized cases, researchers may wish to expose the data in an alternative format. Fortunately, all the data hosted in Open RL Benchmark is accessible through the Weights and Biases API. The following example illustrates how this API can be utilized. From there, researchers can employ any custom script for plotting this data to suit their specific needs. A simple example of such a script is given below, and the corresponding generated plot is shown in Figure \ref{fig:custom_plot}.

\begin{lstlisting}[language=Python]
import matplotlib.pyplot as plt
import wandb

project_name = "sb3"
run_id = "0a1kqgev"

api = wandb.Api()
run = api.run(f"openrlbenchmark/{project_name}/{run_id}")
history = run.history(keys=["global_step", "eval/mean_reward"])
plt.plot(history["global_step"], history["eval/mean_reward"])
plt.title(run.name)
plt.savefig("custom_plot.png")
\end{lstlisting}

\clearpage

\section{Additional Details for the Case Study}
\label{sec:appendix_detailed}

\begin{figure}[ht]
    \centering
    \includegraphics[width=\textwidth]{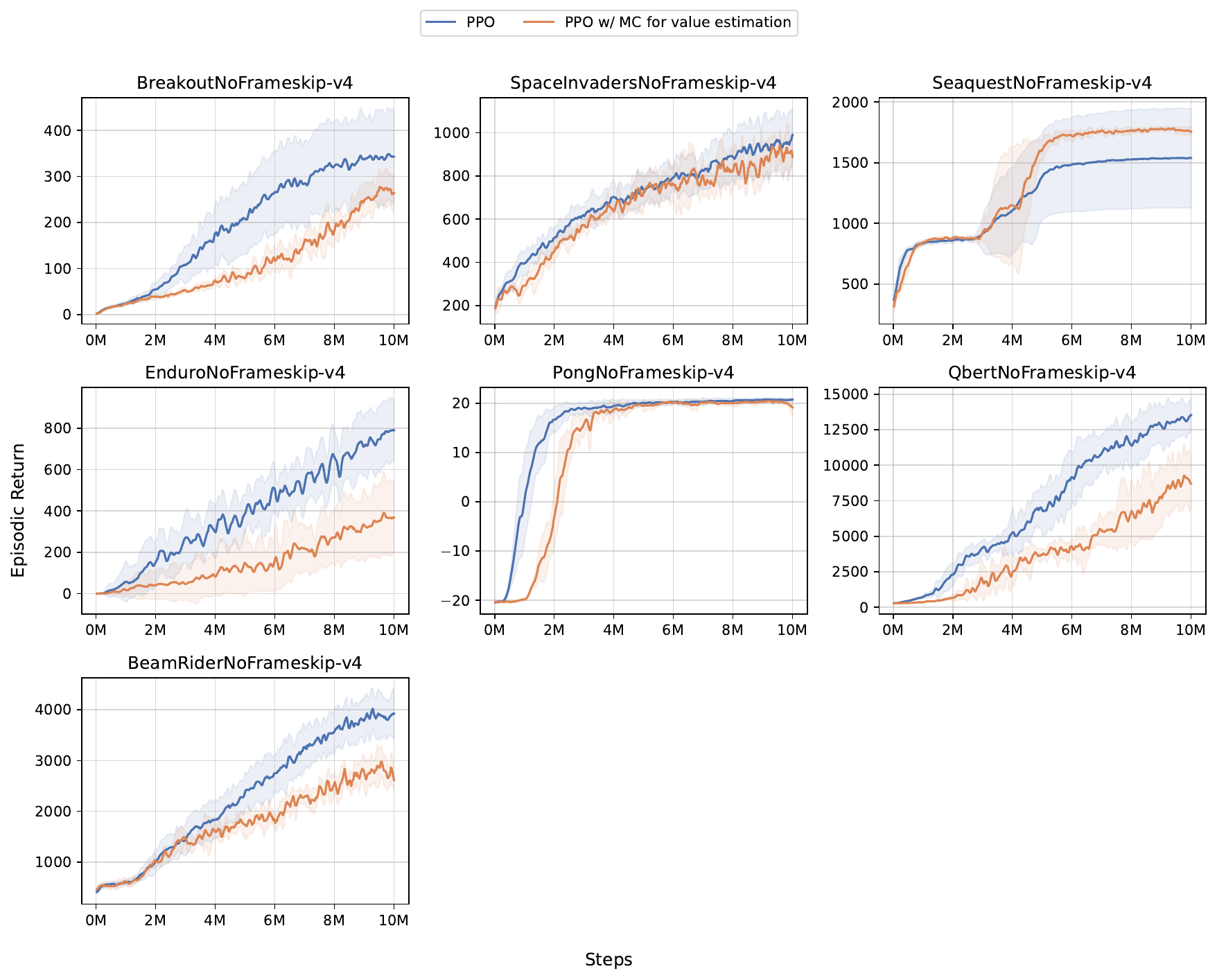}
    \caption{Comparison between the original PPO and the PPO with MC value estimates in various MuJoCo and Box2D environments. Plots represent the evolution of the episodic return as a function of the number of interactions with the environment, and shaded areas represent the standard deviation.}
    \label{fig:gae_for_ppo_value_atari_per_env}
\end{figure}

\begin{figure}[ht]
    \centering
    \includegraphics[width=\textwidth]{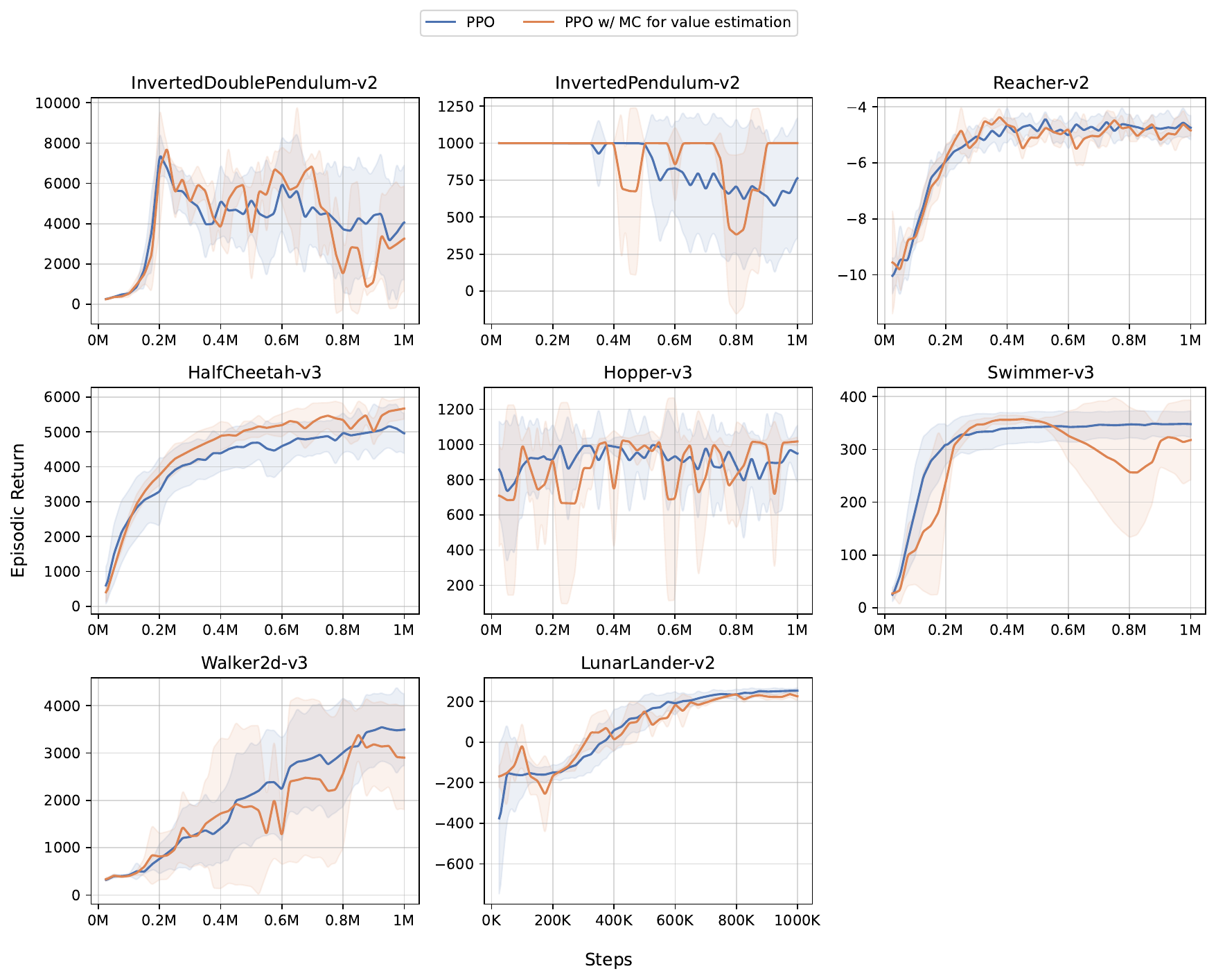}
    \caption{Comparison between the original PPO and the PPO with MC value estimates in various MuJoCo and Box2D environments. Plots represent the evolution of the episodic return as a function of the number of interactions with the environment, and shaded areas represent the standard deviation.}
    \label{fig:gae_for_ppo_value_mujoco_per_env}
\end{figure}

This appendix gives additional results related to the first case study presented in Section \ref{subsec:gae_for_ppo_value}. Figure \ref{fig:gae_for_ppo_value_mujoco_per_env} shows the results by environment for the Atari benchmark, and Figure \ref{fig:gae_for_ppo_value_atari_per_env} shows them for the MuJoCo and Box2d benchmarks. The command lines used to generate these figures are as follows.

\lstset{
  columns=fullflexible,
  breaklines=true,
}
\begin{lstlisting}
python -m openrlbenchmark.rlops \
    --filters '?we=openrlbenchmark&wpn=sb3&ceik=env&cen=algo&metric=eval/mean_reward' 'ppo?cl=PPO' \
    --filters '?we=modanesh&wpn=openrlbenchmark&ceik=env&cen=algo&metric=eval/mean_reward' 'ppo?cl=PPO w/ MC for value estimation' \
    --env-ids BreakoutNoFrameskip-v4 SpaceInvadersNoFrameskip-v4 SeaquestNoFrameskip-v4 EnduroNoFrameskip-v4 PongNoFrameskip-v4 QbertNoFrameskip-v4 BeamRiderNoFrameskip-v4 \
    --no-check-empty-runs \
    --pc.ncols 3 \
    --pc.ncols-legend 2 \
    --rliable \
    --rc.score_normalization_method atari \
    --rc.normalized_score_threshold 8.0 \
    --rc.sample_efficiency_plots \
    --rc.sample_efficiency_and_walltime_efficiency_method Median \
    --rc.performance_profile_plots  \
    --rc.aggregate_metrics_plots  \
    --rc.sample_efficiency_num_bootstrap_reps 1000 \
    --rc.performance_profile_num_bootstrap_reps 1000 \
    --rc.interval_estimates_num_bootstrap_reps 1000 \
    --output-filename static/gae_for_ppo_value_atari_per_env \
    --scan-history \
    --rc.sample_efficiency_figsize 7 4

python -m openrlbenchmark.rlops \
    --filters '?we=openrlbenchmark&wpn=sb3&ceik=env&cen=algo&metric=eval/mean_reward' 'ppo?cl=PPO' \
    --filters '?we=modanesh&wpn=openrlbenchmark&ceik=env&cen=algo&metric=eval/mean_reward' 'ppo?cl=PPO w/ MC for value estimation' \
    --env-ids InvertedDoublePendulum-v2 InvertedPendulum-v2 Reacher-v2 HalfCheetah-v3 Hopper-v3 Swimmer-v3 Walker2d-v3 LunarLander-v2 \
    --no-check-empty-runs \
    --pc.ncols 3 \
    --pc.ncols-legend 2 \
    --rliable \
    --rc.normalized_score_threshold 1.0 \
    --rc.sample_efficiency_plots \
    --rc.sample_efficiency_and_walltime_efficiency_method Median \
    --rc.performance_profile_plots  \
    --rc.aggregate_metrics_plots  \
    --rc.sample_efficiency_num_bootstrap_reps 1000 \
    --rc.performance_profile_num_bootstrap_reps 1000 \
    --rc.interval_estimates_num_bootstrap_reps 1000 \
    --output-filename static/gae_for_ppo_value_mujoco_per_env \
    --scan-history \
    --rc.sample_efficiency_figsize 7 4
\end{lstlisting}

\clearpage
\section{Refine the MuJoCo Benchmark With Stable Baselines3}

\begin{figure}[ht]
    \centering
    \includegraphics[width=\textwidth]{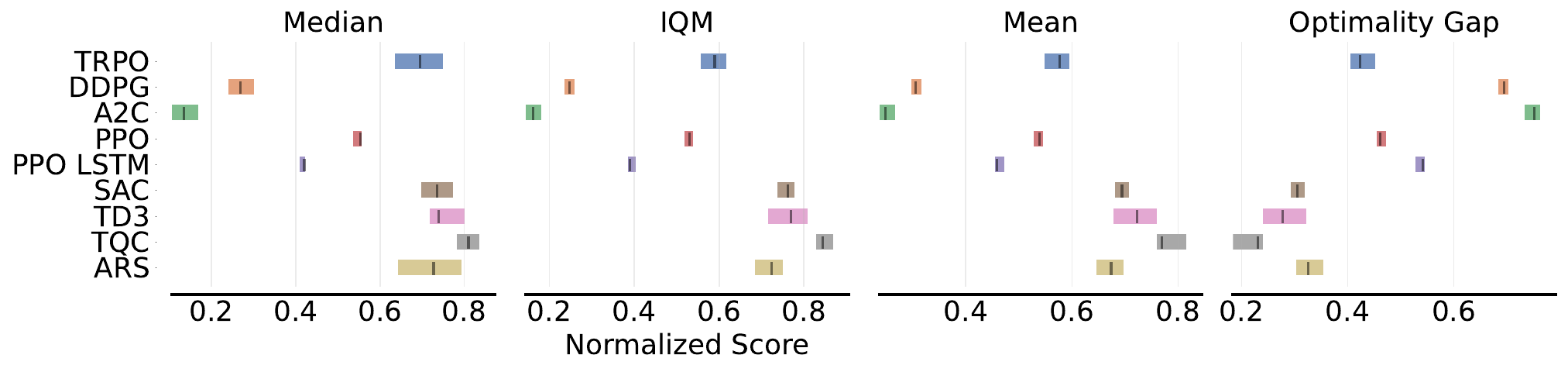}
    \caption{Aggregated final normalized episodic return with 95\% stratified bootstrap CIs on the MuJoCo benchmark of the algorithms integrated into Stable Baselines3.}
    \label{fig:mujoco_agregated_sb3}
\end{figure}

\begin{figure}[ht]
    \centering
    \includegraphics[width=\textwidth]{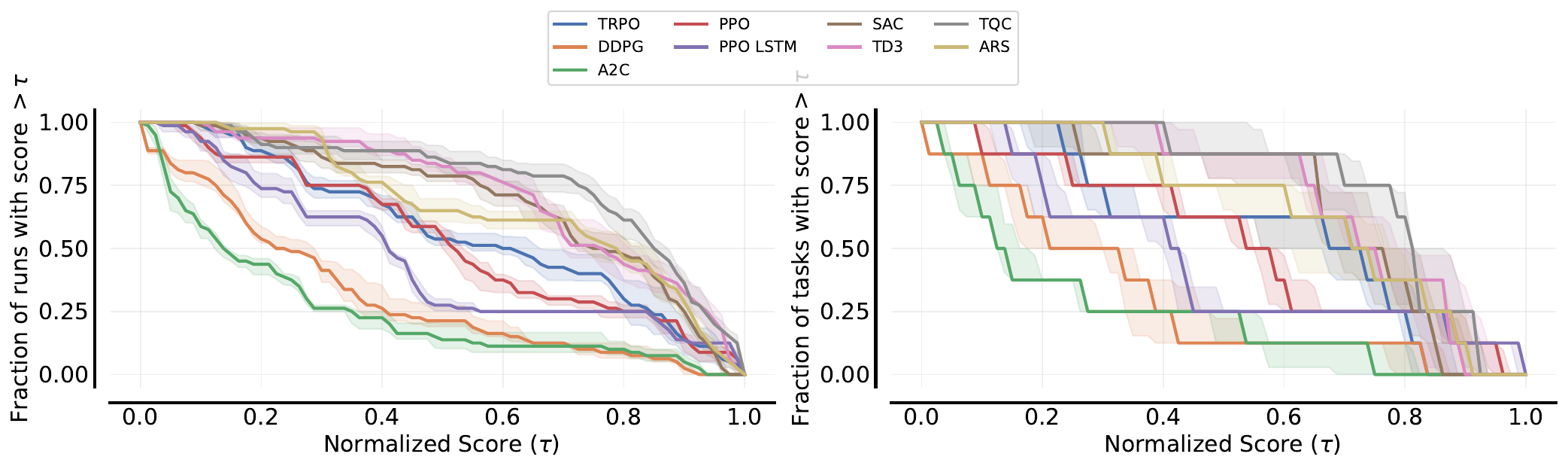}
    \caption{Performance profile of algorithms implemented using Stable Baselines 3 \cite{raffin2021stable} on the MuJoCo benchmark \cite{todorov2012mujoco}. Scores are normalized using the min-max method.}
    \label{fig:mujoco_sb3_performance_profile}
\end{figure}

\begin{figure}[ht]
    \centering
    \includegraphics[width=0.9\textwidth]{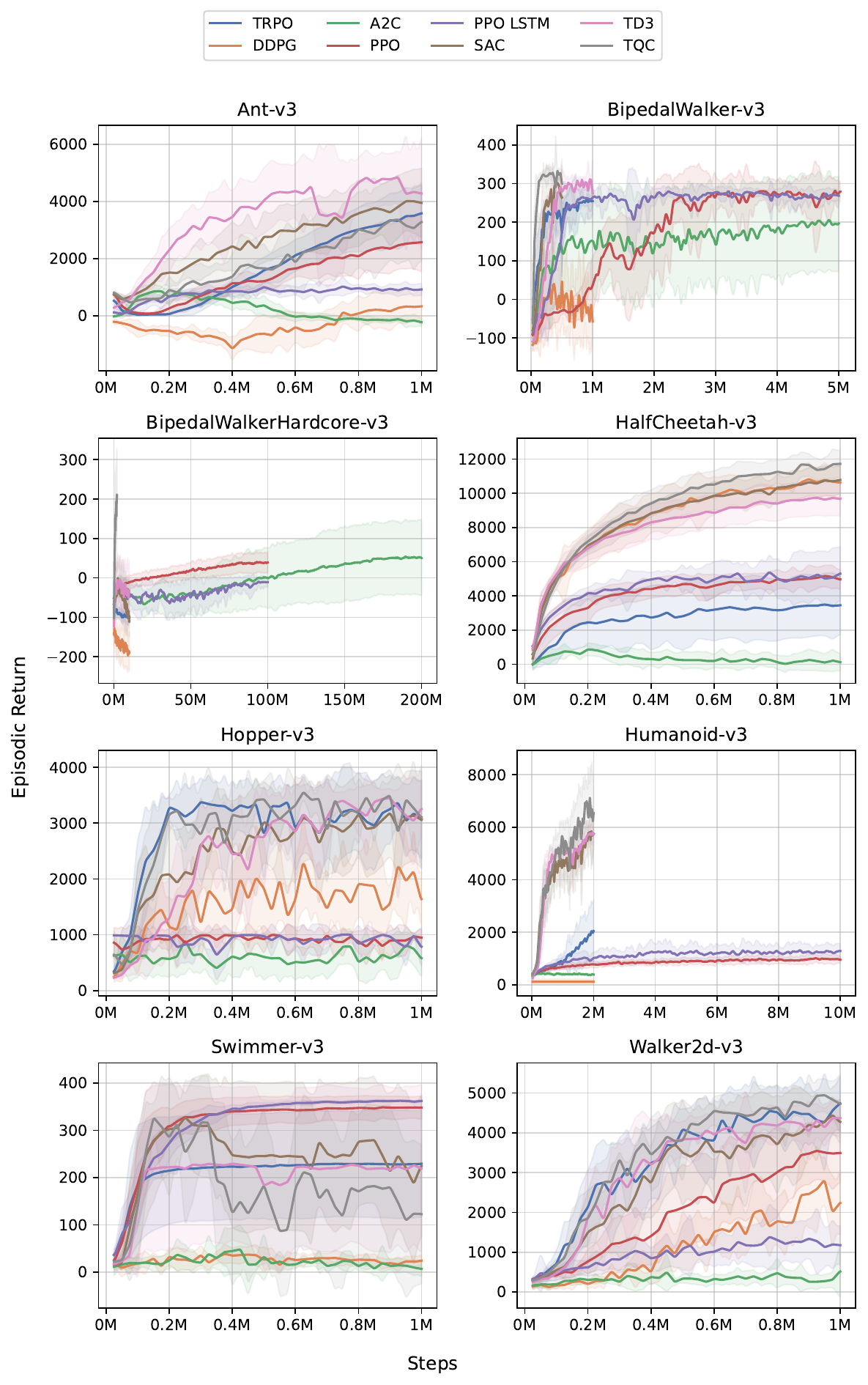}
    \caption{Sample efficiency curves for algorithms on the MuJoCo Benchmark \cite{todorov2012mujoco}. This graph presents the mean episodic return for algorithms implemented using Stable Baselines 3 \cite{raffin2021stable}, averaged across a minimum of 10 runs (refer to Open RL Benchmark for specific run counts). Data points are subsampled to 10,000 and interpolated for clarity. The curves are smoothed using a rolling average with a window size of 100. The shaded regions around each curve indicate the standard deviation.}
    \label{fig:mujoco_curves_sb3}
\end{figure}

In this appendix, we present a syntetic representation of the learning results of the Stable Baselines3 algorithms \cite{raffin2021stable} tested on the MuJoCo benchmark \cite{brockman2016openai, todorov2012mujoco}, whose data is contained in Open RL Benchmark. At the time of writing, data from 757 runs has been used, unevenly distributed between the different experiments. It is important to emphasise that the optimisation of hyperparameters and the training budget vary from one experiment to another.
Consequently, the results should be interpreted with caution. All the hyperparameters and raw data used to generate these curves are available on Open RL Benchmark.
Figure \ref{fig:mujoco_agregated_sb3} shows the aggregation of the final performances following the recommendations of \cite{agarwal2021deep}, and Figure 
\ref{fig:mujoco_sb3_performance_profile} the corresponding performance profiles. Figure \ref{fig:mujoco_curves_sb3} shows the learning curves as a function of the number of interactions.

The command used to generate Figures \ref{fig:mujoco_agregated_sb3}, \ref{fig:mujoco_sb3_performance_profile} and \ref{fig:mujoco_curves_sb3} is as follows\footnote{For Figure \ref{fig:mujoco_curves_sb3}, we're omitting ARS as it was run with many more steps, and its inclusions hinder readability.}.

\lstset{
  columns=fullflexible,
  breaklines=true,
  moredelim=**[is][\color{red}]{@}{@},
}
\begin{lstlisting}
python -m openrlbenchmark.rlops \
    --filters '?we=openrlbenchmark&wpn=sb3&ceik=env&cen=algo&metric=eval/mean_reward' 'trpo?cl=TRPO' \
    --filters '?we=openrlbenchmark&wpn=sb3&ceik=env&cen=algo&metric=eval/mean_reward' 'ddpg?cl=DDPG' \
    --filters '?we=openrlbenchmark&wpn=sb3&ceik=env&cen=algo&metric=eval/mean_reward' 'a2c?cl=A2C' \
    --filters '?we=openrlbenchmark&wpn=sb3&ceik=env&cen=algo&metric=eval/mean_reward' 'ppo?cl=PPO' \
    --filters '?we=openrlbenchmark&wpn=sb3&ceik=env&cen=algo&metric=eval/mean_reward' 'ppo_lstm?cl=PPO LSTM' \
    --filters '?we=openrlbenchmark&wpn=sb3&ceik=env&cen=algo&metric=eval/mean_reward' 'sac?cl=SAC' \
    --filters '?we=openrlbenchmark&wpn=sb3&ceik=env&cen=algo&metric=eval/mean_reward' 'td3?cl=TD3' \
    --filters '?we=openrlbenchmark&wpn=sb3&ceik=env&cen=algo&metric=eval/mean_reward' 'ars?cl=ARS' \
    --filters '?we=openrlbenchmark&wpn=sb3&ceik=env&cen=algo&metric=eval/mean_reward' 'tqc?cl=TQC' \
    --env-ids Ant-v3 BipedalWalker-v3 BipedalWalkerHardcore-v3 HalfCheetah-v3 Hopper-v3 Humanoid-v3 Swimmer-v3 Walker2d-v3 \
    --no-check-empty-runs \
    --pc.ncols 2 \
    --pc.ncols-legend 4 \
    --rliable \
    --rc.normalized_score_threshold 1.0 \
    --output-filename static/mujoco_sb3 \
    --scan-history
\end{lstlisting}

\end{document}